\documentclass[twoside]{article}

\usepackage{theapa}
\usepackage{rawfonts}
\usepackage{jair}

\usepackage{braket}
\usepackage{amsmath}
\usepackage{graphicx}
\usepackage{color}
\usepackage{amsfonts}

\usepackage{comment}

\usepackage{paralist}

\jairheading{72}{2021}{1307-1341}{01/2021}{12/2021}
\ShortHeadings{Quantum Mathematics in Artificial Intelligence}
{Widdows, Kitto, \& Cohen}
\firstpageno{1307}

\usepackage{rawfonts}

\usepackage{braket}
\usepackage{amsmath}
\usepackage{graphicx}
\usepackage{color}
\usepackage{amsfonts}

\newif\ifdraftcolors
\ifdraftcolors
    \newcommand{\dom}[1]{\noindent \textcolor{green}{#1}}
    \newcommand{\tc}[1]{\noindent \textcolor{cyan}{#1}}
\else
    \newcommand{\dom}[1]{#1}
    \newcommand{\tc}[1]{#1}
\fi

\begin{document}

%\begin{frontmatter}

\title{Quantum Mathematics in Artificial Intelligence}

\author{\name Dominic Widdows \email widdows@ionq.com \\
       \addr IonQ, Inc., 4505 Campus Drive, College Park, MD 20740, USA
       \AND
       \name Kirsty Kitto \email kirsty.kitto@uts.edu.au \\
       \addr University of Technology Sydney, PO Box 123, Broadway, NSW 2007, Australia
       \AND
\name Trevor Cohen \email cohenta@uw.edu  \\
       \addr University of Washington, Box 358047, 850 Republican St, Seattle, WA 98109, USA
       }

\maketitle

%\begin{center}
%\em{Preprint: official copy to appear in JAIR (Journal of Artificial Intelligence Research).}
%\end{center}

\begin{abstract}
In the decade since 2010, successes in artificial intelligence have been at the
forefront of computer science and technology, and vector space models 
have solidified a position at the forefront of
artificial intelligence. At the same time, quantum computers have become
much more powerful, and announcements of major advances are frequently in the news.

The mathematical techniques underlying both these areas have more in common
than is sometimes realized. Vector spaces took a position at the axiomatic heart
of quantum mechanics in the 1930s, and this adoption was a key motivation for
the derivation of logic and probability from the linear geometry of vector spaces. 
Quantum interactions between particles are modelled using the tensor product, which
is also used to express objects and operations in artificial neural networks.

This paper describes some of these common mathematical areas, including examples
of how they are used in artificial intelligence (AI), particularly in automated reasoning
and natural language processing (NLP). Techniques discussed include vector spaces,
scalar products, subspaces and implication, orthogonal projection and negation,
dual vectors, density matrices, positive operators, and tensor products.
Application areas include information retrieval, categorization and implication, 
modelling word-senses and disambiguation, inference in knowledge bases, 
decision making, and semantic composition.

Some of these approaches can potentially be implemented on quantum hardware. 
Many of the practical steps in this implementation are in early stages, and some are already realized.
Explaining some of the common mathematical tools  
can help researchers in
both AI and quantum computing further exploit these overlaps, 
recognizing and exploring new directions along the way.

%Quantum logic and
%probability behave differently from their more familiar classical
%formulations: for example, they are based on subspaces, angles and
%projections, rather than sets, measures, and intersections. In some
%cases such quantum alternatives have provided solutions to cognitive
%and information modeling problems that avoid known pitfalls of
%classical set-theoretic approaches and lead to more robust and scalable systems.

%This paper surveys some of these
%areas, focussing on the mathematical techniques at play. In
%several areas, such as vector space logic and composition, 
%the commonalities between quantum theory and AI are a natural 
%consequence of the common use of linear algebra, and the use of quantum theory
%is arguably best seen as a short-cut for bringing historically established
%mathematical techniques into a new application domain. In other cases, particularly 
%modeling uncertainty and decision-making in context, it appears that the use of 
%quantum probability is a direct and largely untapped opportunity.
%We conclude with some suggestions as to which areas we feel hold the most promise for rich streams of %future research, highlighting some of the roads that are less well traveled. 
\end{abstract}

%\end{frontmatter}

%%%
\section{Introduction}
\label{sec:intro}

Vector space models have solidified a position at the forefront of
artificial intelligence (AI), and are a common building block for many kinds of systems. 
From early uses in information retrieval and
image processing, they have spread to semantic modeling in many systems, reaching even greater prominence during the past decade 
due to their use in deep learning. However, vector spaces have a long and illustrious history, and as an analytical technique, vector spaces were established in the mid-1800s. They also provide the foundation for the underlying mathematical model for quantum theory. 

%Specifically, quantum theory uses Hilbert spaces, which are vector spaces defined to have a scalar product which are closed under sequence convergence.
%Importantly, the vector spaces typically used in AI are also Hilbert spaces, which raises an interesting question about how these two domains intersect.

This paper aims to support  readers who are familiar with the use of vectors in AI, and 
interested in mathematical concepts from quantum theory, to make use of this promising set of results and novel techniques in the field of  AI.
Section \ref{sec:vector-intro} summarizes the use of vector spaces in AI and quantum mechanics, including an explanation of how the representation of wavefunctions leads to the study of vector spaces and linear operators in formal quantum mechanics. 
Section \ref{sec:quantum-ir} explains the quantum model for information retrieval, including the use of the quantum logic of subspaces
for modelling concepts and query operations. Section \ref{sec:categories} looks in more detail at models for categories and hypernyms, 
motivating the use of density matrices and positive operators from quantum mechanics. Section \ref{sec:composition} surveys vector models
for semantic composition, in particular the use of tensor products and related operators. Section \ref{sec_implementation} presents
some of the early implementations of these methods on physical quantum hardware, adding fresh vitality to this area of research.

Our hope is that AI practitioners with a grounding in linear algebra will find that
they already know much of the mathematics they need to understand large parts of quantum theory, 
and to begin exploring ways in which they might apply methods inspired by quantum mathematics to the generation of new results in AI.

%We will draw attention to 
%techniques and methods that have already been adapted from quantum theory to the problem of 
%modeling many different scenarios in cognition that are not at present well 
%understood using standard ``classical'' models. We invite AI researchers to 
%participate in an exciting and growing field that has already generated 
%significant interest and clear results, partly in AI but also in related fields
%including information retrieval, machine learning, computational linguistics, psychology, and
%cognitive science.

%%%
\section{Vector Space Models in AI and Quantum Physics}
\label{sec:vector-intro}

Vectors have their roots in the beginning of Euclidean geometry, with the notion that a straight line
can be drawn between any two points and indefinitely extended ({\it{Elements of Geometry}}, Book I, Axioms 1 and 2). 
By the mid-1840s, 3-dimensional vectors and 1-dimensional scalars were thoroughly explored in Hamilton's work on quaternions; 
Cauchy and Hamilton collaborated on the development of matrices and linear algebra; and the algebra of vectors or `extended quantities' 
(German {\it Ausdehnungslehre}) and their products in any number of dimensions is fully-formed in the work of \citeA{grassmann-extension}, 
a summary of which is made available by \citeA{fearnley1979hermann}.

Since then vectors have played a ubiquitous role in scientific advances across a great range of fields,  including 
differential geometry, electromagnetism, relativity, 
machine learning: and of particular interest in this paper, artificial intelligence and quantum theory.

\subsection{Vector Models in AI}

The use of vector spaces in AI goes back at least to information retrieval in the 1960's \cite{switzer1965vector},
the vector model of the SMART retrieval system being a particularly influential example dating from this pioneering period \cite{salton-introduction}. 
Introductions to the vector model for information retrieval are given by 
\citeA[Ch 5]{widdows-geometry} and \citeA[Ch 14]{manning2008introduction}.
In such a system, a document collection is processed into a large table or matrix $M$ whose rows represent terms, whose columns represent documents, and where the entry $M_{ij}$ contains a number measuring the extent to which term $i$ appears in document $j$ (as is depicted in Table \ref{term-doc-matrix-table}).

\begin{table}
\begin{center}
\caption{A term--document matrix}
\label{term-doc-matrix-table}

\vspace{0.1in}

\begin{tabular}{|c|c|c|c|c|c|}
\hline
 & Doc$_1$ & Doc$_2$ & Doc$_3$ & ... & Doc$_m$ \\
\hline
Term$_1$ & $M_{11}$ & $M_{12}$ & $M_{13}$ & ... & $M_{1m}$ \\
\hline
Term$_2$ & $M_{21}$ & $M_{22}$ & $M_{23}$ & ... & $M_{2m}$ \\
\hline
Term$_3$ & $M_{31}$ & $M_{32}$ & $M_{33}$ & ... & $M_{3m}$ \\
\hline
... & ... & ... & ... & ... & ... \\
\hline
Term$_n$ & $M_{n1}$ & $M_{n2}$ & $M_{n3}$ & ... & $M_{nm}$ \\
\hline
\end{tabular}
\end{center}
\end{table}

The rows (or columns) in any such matrix can be added together by adding the corresponding coordinates,
and multiplied by a scalar by multiplying each coordinate, so it is immediately apparent that the rows
(or columns) of any such matrix can be thought of as vectors in a vector space, whose 
dimension is the number of columns (or rows). 

Such a general description 
is naturally available as a model for many other situations that are relevant to AI. Common examples include:

\begin{itemize}
\item The adjacency matrix of any weighted graph \cite[Chapter VIII]{bollobas-modern}. 

\item A collection of grayscale images, each of the same width and height, where 
a row represents an image, columns represent individual pixels, and each coordinate 
represents the shade for that pixel in that image \cite[Ch 3]{geron2019hands}. 

\item A dataset from a medical study where each row corresponds to an individual 
patient and columns correspond to a measurable property or ``vital sign'' 
such as height, weight, age, or blood pressure values.  Many statistical learning techniques begin by assuming their inputs can be modelled in such a fashion \cite{hastie-statml}.

\item Any collection of data projected into a lower-dimensional space, for example, the 
`best fit' subspace produced by a decomposition technique such as Singular Value Decomposition. In the case of textual data,  
the technique of projecting a term-document matrix onto such a subspace is referred to as
Latent Semantic Analysis \cite{landauer-solution}, and today it has many variants, surveyed from an AI point of view by \citeA{turney2010frequency} and from a machine learning point of view by \citeA[Ch. 8]{geron2019hands}. 

\item The activity states of a connectionist network \cite[Definition 2.1]{hinton1990preface,smolensky1990tensor}.
Work on connectionist representations in AI contributed greatly to the development of today's powerful neural network models \cite[Ch 10]{geron2019hands}.

\item The output of a nonlinear training / learning algorithm, for example, the 
feature-weights learned by a neural network. Such models have become increasingly 
prominent in the last few years: techniques that use several intervening layers in the network are often referred to 
as {\it deep learning}, and whether or not a network is deep in this sense, 
the inputs, outputs,
and learned parameters for intervening layers are typically represented as vectors \cite[Ch. 10]{geron2019hands}. 
When applied to textual data the resulting vector models 
are often referred to as {\it word embeddings} in more recent papers since \citeA{mikolov2013efficient}.
\end{itemize}

\noindent
Thus there are many related techniques for deriving collections of vectors from 
empirical observations, and these techniques have become a crucial cornerstone
of AI. Key mathematical benefits include expressing addition of vectors as a pairwise-sum of the corresponding coordinates
(for vectors $x = (x_1, \ldots, x_n)$ and $y = (y_1, \ldots, y_n)$, their vector sum $x+y=(x_1+y_1, \ldots, x_n+y_n)$), 
and measuring similarity (for example, using a scalar product where $x\cdot y = \sum_{i=1}^n x_i y_i$).
For a beginner's introduction to measuring similarity and distance, and the relationship
between scalar products, cosine similarity, and Euclidean distance, see 
\citeA[Ch 4, 5]{widdows-geometry}.

The graded way similarity can be expressed in vector models contrasts with classical Boolean logic:
for example, the introduction of cosine-similarity between query vectors and document vectors as a 
continuous (hence graded) measure of relevance in 
information retrieval enabled systems to return the {\it most relevant} documents ranked in order. This property became vital once document collections became large enough that simply marking documents
as `relevant' or `not relevant' left too many `relevant' documents for a user 
to read.

The use of vector sums for generating graded probabilities in this manner has been termed `superposition' in physics. It enables a combination that is `some of each', 
which contrasts with both Boolean set intersection which is sometimes too specific, and 
Boolean set union which is sometimes too general. Historically, this modeling requirement was recognized very early in the development of quantum theory, where the same vector sum technique is used to 
effectively represent particles that can be measured in more than one way
(this usage is explained in more detail in Section \ref{sec:qm-intro}). 

The adoption of similar methods in AI is more recent, but stems from a similar motivation, as expressed for example
by \citeA{smolensky1986information}:

\begin{quote}
In the subsymbolic paradigm, the semantically interpretable entities are activation patterns, and these combine by \textit{superposition}: activation patterns superimpose upon each other, the way that wave-like structures do in physical systems. This difference is another manifestation of moving the formalization from the discrete to the continuous \ldots
\end{quote}

The core motivation for using vector spaces in AI still has its roots in the benefits
of using continuous representations to complement and sometimes improve upon
discrete Boolean models. These representations can then be put to good use
in problems involving classification \cite[Ch 14]{manning2008introduction},
\cite[Ch 3-5]{geron2019hands} and logical reasoning \cite{widdows2015reasoning}.
Machine learning techniques for building such representations have improved considerably,
with the use of deep neural networks, sequence modelling, transformers and attention-based techniques
(for a survey, see \citeA[Ch 16]{geron2019hands}). 
Vector models are used throughout AI and related fields,
all the way from pioneering work in information retrieval to today's neural network language models. So it makes sense to ask whether vector methods 
derived from other areas of science are useful for learning and reasoning with such models.

\subsection{Why Quantum Theory?}

Quantum physics appears often and prominently in the science and technology news.
Two popular quantum physics topics are immediately relevant to AI: firstly, 
the suggestion that natural intelligence may 
involve quantum physics directly, put forth most famously by \citeA{penrose1999emperor}; secondly,
the increasing technological progress being made in quantum computing, described by e.g., \citeA{reiffel-quantumcomputing,bernhardt2019quantum}. 
The popular appeal of quantum physics and computing is easy to appreciate: it challenges established norms; 
it is at forefront of innovation; it is real yet mysterious \cite{aaronson2013quantum}. 

It is commonly said that quantum theory applies only to particularly small physical systems,
because many of the discrete quantization and interference effects explained by quantum mechanics
are not significantly observable at large scale.
However, the mathematics developed as part of vector models in 
quantum theory can also be used to describe larger human-scale phenomena, such as individuals or 
groups making decisions or searching for information. 
One of the pioneers and leaders in this area has been the physicist Diederik Aerts (see \citeA{aerts1993quantum} and onwards).
The techniques and models proposed by Aerts and others are sometimes called 
quantum-inspired, quantum-like, or generalized quantum models \cite{khrennikov-ubiquitous}. 
This makes a case that quantum {\it mathematics} is worth investigating in 
situations outside the traditional microscopic domain of quantum {\it physics}, without
answering the challenging ontological questions that arise when asking how quantum physics might
affect larger systems. In this context, the main argument in favor of quantum mathematical models is utilitarian:
many solutions to known problems have been improved in a number of different ways. Books and papers discussing the 
application of quantum mathematics are available in several fields including: information retrieval  \cite{melucci2015introduction}; machine learning  \cite{wittek-qml}; cognitive science \cite{busemeyer.bruza:quantum}; and economics 
\cite{khrennikov-ubiquitous}. 

Within AI itself, the book by \citeA{wichert:principles} provides a graduate-level text on quantum computing 
and AI, grounded in computational theory, physics, and quantum algorithms. In contrast, this paper 
provides a much more thorough introduction to tensors and their uses. 
Potential quantum approaches in AI are well-surveyed in the article by \citeA{ying:quantum}, which gives a thorough
introduction to qubits and quantum gates. In comparison,  our work here focusses upon two decade's hindsight and progress in the use
of vector models and related mathematics including subspaces, projections, and tensor products which has flourished in AI.

Simultaneously, the increasing power of quantum computers has turned quantum mechanics from a fascinating mystery
to an increasingly practical opportunity, and quantum advantages such as
the promise of exponential memory savings are bound to attract the attention
of researchers in AI. Some recent advances in AI and quantum computing
are presented in Section \ref{sec_implementation}.

A question remains: \emph{Why} does the mathematics of quantum theory appear to support these types of scenarios? Many different explanations are possible depending upon the philosophical stance that one adopts, but for now we will remain agnostic. We will return to this important question in Section~\ref{sec:again-why}, with more insights from the mathematics of quantum theory to support our proposed answer to this question. For now we will continue with a brief introduction to the topic of quantum theory itself.

%%%
\subsection{Formal Quantum Mechanics}
\label{sec:qm-intro}

To appreciate the appeal of quantum-inspired techniques in vector space models we must work to bridge a gap between the physical results
of quantum theory (as predicted by quantum mechanics), and its mathematical formulation 
(in terms of state vectors 
and operators on Hilbert space). 
 In the process, we hope to show that topics considered to be philosophically 
challenging in quantum physics, such as non-locality, quantum disjunction, and entanglement,
are mathematically quite straightforward. Many of them rely upon simple operations such as projecting 
lines onto planes, finding the plane spanned by two lines, and multiplying vectors to generate square matrices which can be added together. 
The mathematical formulation of 
quantum mechanics mainly uses standard linear algebra, which means that many of its techniques are already familiar to those working in AI. An excellent introduction is provided by  \citeA{isham:quantum} and of course \citeA{dirac-quantum}. Of particular note, the formalism of quantum theory requires us to move from a model of reality that assumes localised particles, to one that assumes waves to be fundamental, and these can be represented as vectors using Fourier series. 

Quantum theory was invented to describe the behavior of subatomic particles, and has successfully explained
and predicted many behaviors that could not be adequately accounted for by late 19th century
classical conceptualizations. These classical models understood the universe as made up of 
particles, imagined to consist of Euclidean `points' (that is, things that have no spatial
extent of their own, and can be located anywhere in a continuous space), and vibrating waves such as light.\footnote{
This is an accepted if a somewhat simplified view of physics before 1900. 
For example, Sir Isaac Newton postulated the existence of distinct corpuscles of light,
and both Democritus and Plato postulated the existence of particles with some sort of shape 
or spatial extent.}
In the early decades of the 20th century, particle-like behavior was observed in light using experiments that 
increased our understanding of the polarization and interference of photons \cite[Ch 1]{dirac-quantum}. 
More experiments revealed that matter exhibited wave-like as well as particle-like behaviour.
A new theory was needed, one in which both
`matter' such as atoms, electrons, and protons, and `radiation' such as light exhibited both 
wave-like properties (such as interference) and particle-like properties (such as coming in 
discrete chunks). 

\subsection{Waves as Vectors Using Fourier Series}
\label{sec:waves}

Mathematically, any wave can be represented as a function. For example, a standing wave on a string of length $L$ is
represented as a function $f:x \in [0, L] \rightarrow \mathbb{R}$, and a dynamic (moving) wave is represented as a function of two variables $f: (x\in [0, L], t \in \mathbb{R)} \rightarrow \mathbb{R}$.
Real functions can be
multiplied by any real number, and added together pairwise, and with these natural definitions, 
the standard vector space axioms
are satisfied: thus the set of all such real functions forms a vector space. Functions to the reals are vectors.

\begin{figure}
    \centering
    \includegraphics[width=\linewidth]{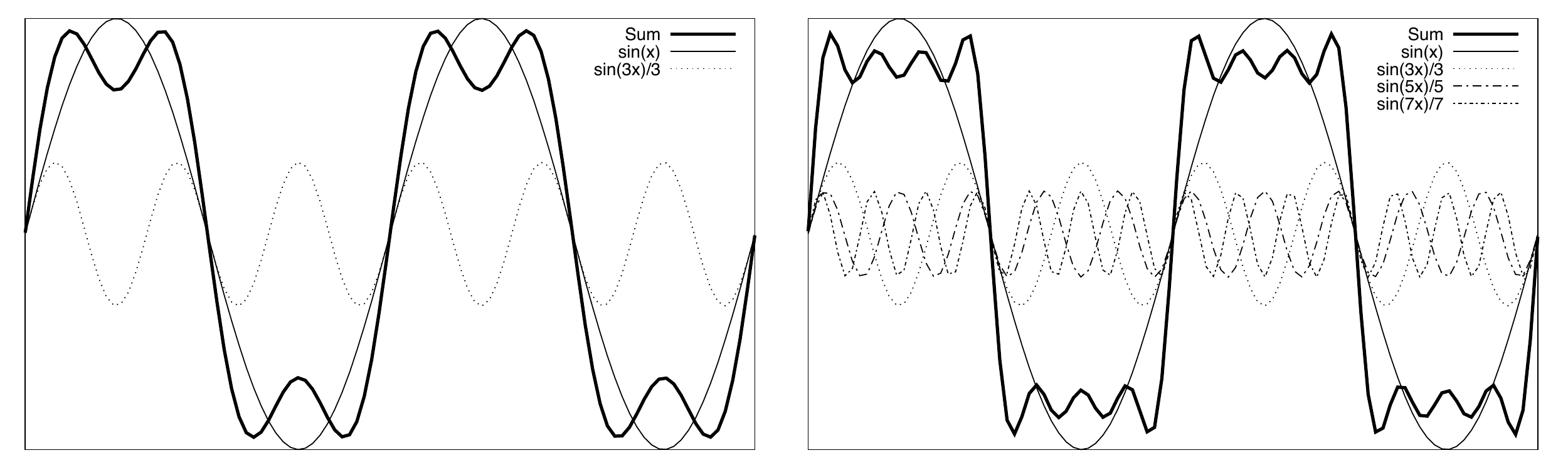}
    \caption{Fourier series approximations for a square wave, showing contributions from 
    the first two and a the first four nonzero harmonics.}
    \label{fig:fourier_square}
\end{figure}

Moreover, piecewise smooth functions of one variable can be represented as weighted sums
of their basic constituents. Such representations were introduced by Joseph Fourier (1768--1830),
who demonstrated that under certain continuity conditions a real function $f$
over the interval $[0, 2\pi]$ can be uniquely represented as a sum of harmonic functions,
\[
    f(x) = \sum_{0}^{\infty}a_k \sin(kx) + b_k \cos(kx). 
\]
This sum is called the Fourier series of $f$, and the $a_k$ and $b_k$ are called
its Fourier coefficients. \dom{The example in Figure \ref{fig:fourier_square} shows
the first few harmonic functions that sum up to form closer and closer approximations to a square wave,
the expansion in this case being $\sum_{n=1,3,5,\ldots}\sin(nx)/n$.}

In vector space terms,\footnote{Introductions to vector spaces --- 
basis vectors, coordinates, dimensions, and linear dependence --- can be found in 
texts on linear algebra \cite{strang1993introduction}, quantum computing \cite[Ch 2]{nielsen2002quantum}, 
natural language processing \cite[Ch 5]{widdows-geometry}, and in many online tutorials.}
 the sine and cosine functions form a
basis for the vector space of real functions, and the coefficients $a_k$, $b_k$ 
are the coordinates of the vector $f$ in this basis. 
\dom{For example, in the case of the square wave of Figure \ref{fig:fourier_square},
the $a_k$ coordinates are $(0, 1, 0, \frac{1}{3}, 0, \frac{1}{5}, 0, \frac{1}{7}, \ldots)$ 
and the $b_k$ coordinates are all zero.} 
To represent an arbitrary function with
perfect precision, an infinite number of such coordinates may be needed, so 
the vector space is infinite dimensional in principle. Infinite dimensions are needed in quantum {\it mechanics}
to represent potential values of the position and momentum operators \cite[Ch 2]{khrennikov-ubiquitous}, 
but in quantum {\it computing}, 
the most typical vector space building block is the 2-dimensional complex vector
space $\mathbb{C}^2$ which is used to represent the state of a single qubit \cite[\S 1.2]{nielsen2002quantum} \cite[\S2]{ying:quantum}.

The vector space of piecewise smooth functions has a natural scalar product: 
for two real functions $f$ and $g$ defined over the interval $[a, b]$, 
we define their scalar product to be $\int_a^b f(x)g(x)\, dx$. (This scalar product is often called 
the inner product and is typically written $\langle f, g \rangle$ in functional \linebreak analysis.)
Crucially for the Fourier theory, for all natural numbers $n$, $m$, the inner product \linebreak
$\int_0^{2\pi} \sin(nx) \sin(mx)\, dx$ evaluates to 0, unless $n = m$ in which case it
evaluates to 1. The same result holds for products of cosine functions, and all products of 
a sine and a cosine function. Thus these sine and cosine functions form not only a basis, but an
orthonormal basis for the vector space of smooth real functions. (An orthonormal basis is one
where each basic vector has unit length and each pair of distinct basis vectors are orthogonal to one another. Note that 
in machine learning, `one-hot encodings' form a  kind of orthonormal basis \cite[Ch 2]{geron2019hands}.

Representing functions using coordinates via Fourier series expansion is only one option. Many other such expansions were later found, for single-variable functions
over the real numbers, and also for functions over other spaces such as the sphere, the so-called spherical
harmonic functions. The basis of spherical harmonic functions is crucial to the solution
of the Schr\"odinger equation for the hydrogen atom, and in fact each possible energy level
for an electron corresponds to a particular spherical harmonic function. Because of this, the
number and distribution of spherical harmonic functions accounts directly for the structure
of the periodic table of elements in chemistry \cite[\S39]{dirac-quantum}, but they are not just limited to physical systems. Interestingly, this mathematical formalism has been fruitfully used in the rendering of images, a point that was first recognised by \citeA{sloan.jan.ea:precomputed}. Spherical harmonics can be seen as a way to encode information over a sphere, in general, beyond their historical origins in solving partial differential equations in mathematical physics. 

By the end of the 19th century this rich mathematical theory was fully mature, before 
quantum physics was developed. Functions, series,
limits, conditions for convergence, and the uses of different bases of functions to
represent different physical systems (typically arising from solutions to partial 
differential equations) were well-understood. Around this time David Hilbert added to the earlier vector spaces of Hermann  \citeA{grassmann-extension}, 
working specifically with vector spaces
with a well-defined scalar product and where every convergent sequence has a limit.
Such a space is today called a Hilbert space. 

Later, Paul \citeA{dirac-quantum} introduced the bra-ket notation, designed especially for
calculations involving wavefunctions of quantum systems represented as vectors in such a Hilbert space. 
In Dirac notation,
the wavefunction $\psi$ is written as a `ket' vector $\ket{\psi}$, and then the scalar product
of $\ket{\phi}$ and $\ket{\psi}$ is written as $\braket{\phi|\psi}$. The object $\bra{\phi}$
in this expression has its own interpretation as a {\it covector} or dual vector, called a `bra' 
vector in Dirac notation: for any vector space $V$ over the field $F$, the dual space $V^*$
is the space of $F$-linear mappings from $V$ to $F$, whereupon it is easy to show that 
$V^*$ is isomorphic to $V$, and a particular scalar product (written as a bilinear form $\langle u, v\rangle$)
induces one such isomorphism by the identity $u \rightarrow \langle u, \_ \rangle$ \cite[\S 2.1.4]{nielsen2002quantum}.
One of the key reasons for using Hilbert space as the setting for quantum mechanics is that the existence
of such a scalar product makes it possible to switch between the ket vector $\ket{\psi}$ and the
bra covector $\bra{\psi}$ whenever this is convenient, and the similarity between the angle bracket
notation for bilinear forms and the bra-ket notation makes this elegantly obvious. 
Introductions to 
Dirac notation are given by  \citeA[Ch 2]{nielsen2002quantum}, \citeA[p. 104]{rijsbergen-geometry}, 
\citeA[Ch 2]{bernhardt2019quantum}, and 
the original presentation in Dirac \citeyear{dirac-quantum} is still very readable.

\begin{figure}
\includegraphics[width=\linewidth]{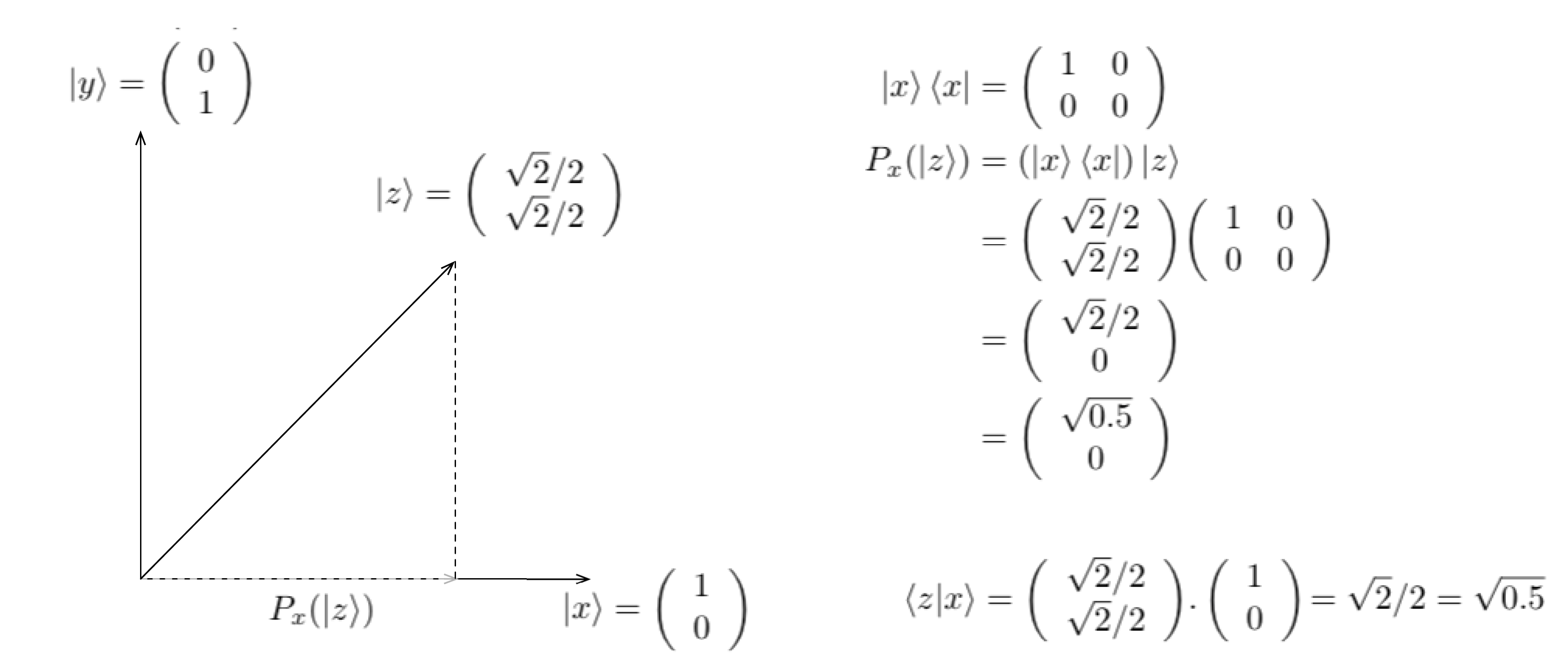}
\caption{Projection using the outer product of the x axis with itself, $\ket{x}\bra{x}$.}
\label{fig:projections}
\end{figure}

\tc{Figure \ref{fig:projections} shows the projection of a ket vector $\ket{z}$ onto the $x$-axis whose ket vector is $\ket{x}$.
The projection operator is constructed as $\ket{x}\bra{x}$, which represents the $\textit{outer product}$ of the vector 
$\ket{x}$ with itself. For $\ket{x}$, this outer product is a diagonal matrix with zero values except for a 1 in the top 
leftmost quadrant. Multiplying $\ket{z}$ by this matrix results in its projection onto the x axis, $P_x(\ket{z})$. Of note, the 
length of this projected vector is $\sqrt{2}/2$, which is also the scalar product between $\ket{z}$ and $\ket{x}$, or 
$\braket{z|x}$ (as well as the cosine of the angle between these unit length vectors). This projection can be interpreted 
probabilistically --- the length of the projection on $\ket{x}$ gives the \textit{probability amplitude} of observing the 
superposition $z$ in state ${x}$. Squaring this probability amplitude gives the probability itself, in this case 0.5.  }

It should also be noted that elements of the dual space $V^*$ are linear operators on $V$
{\it and} vectors in their own right: and elements of $V$ are also linear operators on $V^*$.
Matrices present another such example. An $m\times n$ real matrix can be used to represent a linear mapping from 
$\mathbb{R}^n$ to $\mathbb{R}^m$ using matrix multiplication. At the same time, it is easy to see that
that matrix addition adds the coordinates of two $m\times n$ matrices just like vector addition, and that the set of all 
$m\times n$ is also a vector space --- matrices are both vectors and operators on vectors. 

\dom{Another crucial definition related to dual vectors and linear operators is that of the {\it adjoint} operator. The adjoint
of a linear operator $A$ is the operator $A^\dagger$ such that $\braket{A^\dagger (\phi)|\psi} = \braket{\phi|A (\psi)}$
for all $\ket{\phi}$, $\ket{\psi}$. The definition of a Hilbert space guarantees that adjoints
exist and are unique, and that $(A^\dagger)^\dagger = A$ \cite[\S2.1.6]{nielsen2002quantum}.}

\subsection{Observation and Uncertainty}
\label{sec:observation}

Experimental measurements on quantum systems are represented in quantum theory using an \emph{observable}, which is a linear operator $A$ that acts on the vector $\ket{\phi}$
to give a vector $\ket{A\phi}$. If this observation is performed twice in quick succession, the
same state is always observed, so $\ket{A\phi}$ is invariant under the action of $A$. 
Such a situation occurs mathematically for states represented by eigenvectors of the
operator $A$, which are those vectors for which $A\ket{\psi_k} = \lambda_k\ket{\psi_k}$ (i.e. vectors that do not change their orientation when the operator is applied), where $\lambda_k$ is
the $k^\mathrm{th}$ eigenvalue of $A$. 
The probability of observing the system $\ket{\phi}$ in the 
eigenstate $\ket{\psi_k}$ is given empirically by $|\braket{\phi|\psi_k}|^2$. As well as being  
probability amplitudes, in geometric terms the scalar products $\{\braket{\phi|\psi_i}\}$ are also the coordinates for the 
vector $\ket{\phi}$ when it is expanded in the basis given by the eigenvectors $\{\ket{\psi_i}\}$.
So to understand the way $A$ operates on $\ket{\phi}$, it is enough to write the vector $\ket{\phi}$
as an expansion $\sum \lambda_i \ket{\psi_i}$, where $\lambda_i = \braket{\phi|\psi_i}$. Such an
expansion works only if the operator $A$ has a suitable spectrum of eigenvectors with real eigenvalues
(they must be real if their squares are to be interpreted as probabilities). It follows from linear
algebra that $A$ must be self-adjoint, that is, $A=A^\dagger$ or more explicitly, 
$\braket{A\psi|\phi} = \braket{\psi|A\phi}$ for all $\ket{\phi}$, $\ket{\psi}$.
\dom{In terms of coordinates and matrices, an operator is self-adjoint if and only if it is represented by a 
matrix that is equal to its conjugate transpose: that is, $a_{ji} = \bar{a}_{ij}$ for all $i, j$.
Such a matrix is called {\it Hermitian}.}

Other results in quantum mechanics can now be derived relatively easily. For example, a \emph{quantum superposition} arises whenever two (or more) state vectors are added together, a scenario that commonly occurs in many vector space AI models. 
Similarly, the Heisenberg Uncertainty Principle follows from the fact that linear operators do not usually
commute with one another (as is well-known from matrix multiplication, $AB$ and $BA$ are 
not usually the same), and the size of the uncertainty is bounded by the commutator $AB - BA$.
This follows directly from the mathematical representation, which as we have seen, follows from the
basic requirement that the system be represented in a way that is amenable to linear superposition
(hence the use of vectors), and that observations be represented in such a way that performing the
same observation twice in succession gives the same answer. 

Understanding the relationship between such mathematical 
implications and behaviors in the physical world is one of the key areas of
research in quantum foundations. 
For example, on the Uncertainty Principle itself, 
\citeA[p. 114]{rijsbergen-geometry} 
pointed out ``It is surprising that such a famous principle in physics is implied 
by the choice of mathematical representation for state and observable in Hilbert space.'' 
While the philosophical status of this implication may still be debated, the 
claim that a change of basis leads to different and unpredictable measurements has gone
from a paradox in the 1930s, to the design of encryption protocols such as BB84 in the 1980s 
\cite[Ch 3]{bernhardt2019quantum}, to commercial offerings of quantum key distribution today.
Other properties such as nonlocality and entanglement
will be introduced and discussed mathematically in this paper: as described by
authors including \citeA{coecke2017picturing}, although some of these mathematical consequences were first perceived as awkward difficulties for realistic physics,
they have now become cornerstones of how quantum computers work in practice. 

%%
% Unnecessary repetition?
%\subsection{Opportunities for Quantum Models in AI}
%
%Presentations of quantum mathematics like the above often leave readers unimpressed. Instead of talking
%about subatomic particles and wave interference, the discussion is about vectors and operators.
%At this abstract level, the question ceases to be ``What could quantum theory have to do with AI?''
%and instead becomes ``If it's just that they both use vectors, why could that be interesting?''
%
%The first reason is theoretical and mathematical.
%The way that quantum theory uses vectors spaces introduces a logic, a probability theory, and composition
%operators that are in some cases
%better-suited for representing cognitive and information processing operations. So it is not just
%that quantum theory and AI share the use of vectors as a spatial or geometric model: the opportunity is 
%that AI can benefit from the logic and probability that follows from the geometry of quantum theory.
%The second and more recent reason is that advances in hardware are such that 
%actual quantum models for AI have begun to be implemented on quantum computers,
%with experiments such as those of \cite{wang2020quantum} and \cite{fischbacher2020single} for image classification and
%\cite{meichanetzidis2020qnlp} for natural language processing (NLP). 
%
%In what follows we shall consider the manner in which these advantages can be used to good effect
%in core areas of AI, with examples drawn particularly from information retrieval and NLP.

%%%
\section{The Quantum Formulation of Information Retrieval}
\label{sec:quantum-ir}

Information retrieval was pioneering in the use of vector spaces for representing language and information, 
and the first area related to AI and computational linguistics to
be described thoroughly from a quantum theoretical point of view (by \citeA{rijsbergen-geometry}).
Because of this head start, information retrieval and related word vector representations are a 
prominent source of examples in this paper.
This section introduces these applications of vectors and quantum theory
to information retrieval, explaining the way that cosine similarity can be seen as a projection operator, and how
projection operators have been used to model other logical operations including conditionals, negation and disjunction in the logic of subspaces and projections which is still
called `quantum logic' following \citeA{birkhoff-logic}.

The vector space model for information retrieval (summarized by \citeA{salton-introduction} and \citeA[Ch 6]{manning2008introduction})
stands out as one of the first AI models that was given a distinctly quantum theoretical formulation, developed by \citeA{rijsbergen-geometry} and \citeA[Ch 7]{widdows-geometry}. Van Rijsbergen realized that the Hilbert-space formulation of quantum mechanics and the vector space model for information retrieval have much in common, and are in many ways identical. This paved the way for much work in understanding and exploiting commonalities between quantum mechanics and information retrieval. A thorough summary of progress in the subsequent decade is given by  \citeA{melucci2015introduction}.

Van Rijsbergen \citeyear{rijsbergen-geometry} began to join these areas using the simple observation that if we use the Euclidean norm to measure distance, then the vector spaces used in information retrieval are all trivially Hilbert spaces (since they are finite-dimensional, they are also complete).
Typically, vector-based information retrieval systems use the cosine measure to give similarities between queries and documents, which gives the same ranking as using Euclidean distance with normalized vectors (as explained in detail by \citeA[\S5.5]{widdows-geometry}), so while the strictly metric properties of Euclidean space are often not emphasized when calculating similarity, they are present mathematically. Given a query vector $q$ and a document vector $d$ (both of unit length), the cosine similarity $q \cdot d$ is also the length of the projection of $d$ onto $q$. Using Dirac's bra-ket notation, this can easily be written as $\braket{q|d}$.

This theory was then developed using the non-commutative behavior of projection operators to model the interaction of a document
being relevant to a particular query, and about a particular topic. It can be exploited to model
conditional logic and implication for IR in vector spaces \cite[Ch 5]{rijsbergen-geometry}. 
Thus there are ways to adapt conditional operators to vector space models. Importantly,  much of the work involved can proceed (and historically has proceeded) quite independently of quantum mechanics. 
Powerful logical operations in Hilbert spaces are available, and whether or not they are inherently `quantum' is partly a historical question. Here, we will argue that they are useful either way, but we will return to this this question in Section~\ref{sec:again-why}.

\subsection{An Intuitive Primer: Quantum Logic as Vector Subspaces and Projections}
\label{sec:subspacesQL}

Recognizing that similarity is implemented as a kind of projection operator takes the parallel between QM and IR further, because the observables in a quantum system are often represented as projection operators in a Hilbert space. The logical structure of such projections was discovered and analyzed in a seminal paper called {\it The Logic of Quantum Mechanics} \cite{birkhoff-logic}, and has since been called just `quantum logic', or sometimes the `standard logic' on a Hilbert space \cite[Ch 1]{varadarajan-quantum}. Geometrically speaking, the key to understanding the contrast between quantum logic and Boolean logic is to see that Boolean logic is modelled by {\it subsets}, whereas quantum logic is modelled by {\it subspaces}, that is, subsets that are themselves vector spaces.
For example, in 3-dimensional space, any arbitrary collection of points can be considered a subset, but only the lines and planes are subspaces. Note also that with a scalar product, each subspace $P$ can also be used to define the operation of {\it orthogonal projection}
onto that subspace, and if $\ket{p_1}, \ldots, \ket{p_n}$ is an orthonormal basis for $P$, this projection takes the particularly simple
form $\pi_P = \sum_{i=1}^n \ket{p_i}\bra{p_i}$ \cite[\S 2.1.6]{nielsen2002quantum}. 
The `similarity' between a vector $\ket{a}$ and the subspace $P$ can then be defined as the 
magnitude of $\pi_P(\ket{v})$, and when $P$ is one-dimensional and $\ket{a}$ has length one, this recovers the familiar definition of cosine similarity between two vectors.

\begin{figure}
    \centering
    \includegraphics[width=\linewidth]{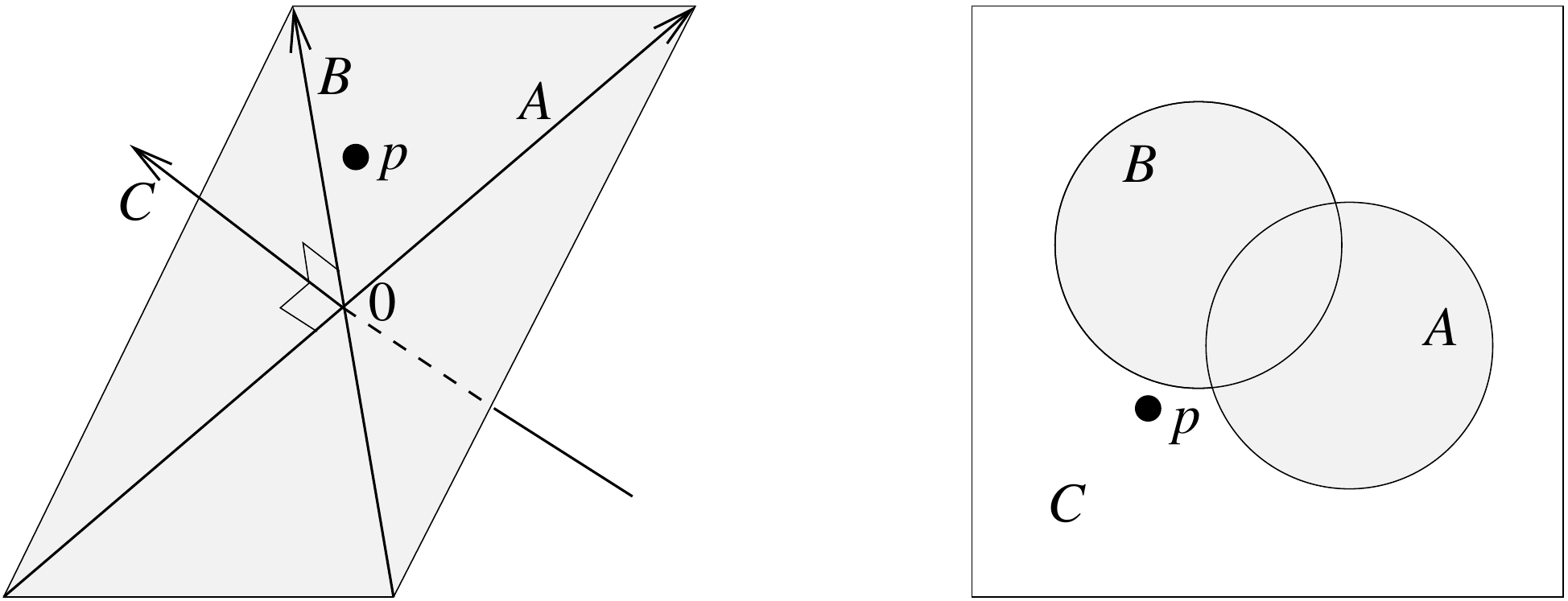}
    
    \caption{Vector Subspaces contrasted with Venn Diagrams of Sets. On the left, a Euclidean picture of the quantum logic of subspaces, and on the right, a Venn diagram picture of Boolean logic of sets.}
    \label{fig:subspaces}
\end{figure}

A sketch highlighting some of the differences between Boolean and quantum logic is given in Figure \ref{fig:subspaces}. In the realm of quantum logic (depicted left), we can see that if a point
$p$ can be written as $\lambda A + \mu B$ for some $\lambda$ and $\mu$, it is considered part of the disjunction $A\vee B$ even if
$p$ is itself in neither $A$ or $B$. The negation $C$ is not the entire rest of the space, but is strictly the {\it orthogonal} complement $C^\perp$.  
This is a big contrast with Boolean logic --- in the familiar Venn diagram on the right hand side of Figure \ref{fig:subspaces}, if the point $p$ is in neither $A$ nor $B$, then by definition it is part of $C=\neg(A\vee B)$. In quantum logic, if $p = \lambda A + \mu B + \nu C$ and none of these coefficients are zero, then we cannot say either that $p\in A\vee B$ or that $p\in C = \neg(A\vee B)$, but instead $p$ has some of each. The conjunction operator behaves the same way in both quantum and classical logic (because the intersection of two subspaces is itself a subspace, for example, the intersection of two planes is typically a line), but disjunction and negation behave very differently.
Finally, in the quantum scenario we find that if the set ${A, B, C}$ forms an orthonormal  basis, then
the coefficients $\lambda$, $\mu$, and $\nu$ correspond
precisely to the {\it probability amplitudes} for observing $p$ in one of these basis states in quantum mechanics \cite[Ch 2]{isham:quantum}.

\subsubsection{Subspace Disjunction}

In quantum logic, the disjunction of two one-dimensional subspaces (lines through the origin) is the plane that spans them both.
In Figure \ref{fig:subspaces} (left), the example is that the disjunction of the lines $OA$ and $OB$ is the plane $OAB$, and the notation is usually
simplified by leaving out the origin $O$ and saying that the disjunction of the lines $A$ and $B$ is the plane $A\vee B$.
Note that this includes points that are not in either of the initial lines (these points correspond to a superposition in quantum mechanics.) In logical terms, we say that quantum logic is {\it non-distributive}, because it fails to satisfy the distributive law $P \wedge (Q \vee R) = (P\vee Q) \wedge (P\vee R)$. 

More intuitively, the quantum disjunction 
interpolates (and sometimes extrapolates) from its inputs. A conversational example that contrasts such inclusion might be:

\begin{quote}It's 70 or 80 miles from Oxford to Birmingham and buses leave at 9am or 11am.\end{quote}
\noindent
Without even thinking about it, an adult reader will assume that the actual distance (about 74 miles) is expressed perfectly well by the phrase ``70 or 80 miles'',
but will not assume that there is also a 10am bus departure. 
%In the first case, the disjunction is more quantum, and in the latter, more Boolean.
\dom{In the first case, the combination is continuous, whereas in the second case, it is discrete, and while Boolean logic gives a better model
for the latter, the former is more effectively modelled by quantum disjunction. (This is just an illustrative example, and 
several other approaches such as fuzzy logic can be motivated 
in a similar way.)}
The notion that 
an object can be modelled in a space that combines $A$ and $B$ without it being part of either $A$ or part of $B$ is a feature of quantum systems
and their non-separability that we will encounter again later (and more strikingly) with tensor product representations and quantum entanglement.
With subspaces, this affects the way that categories or natural kinds might be modelled in an IR system, 
a point discussed more fully by \citeA[Ch. 2]{rijsbergen-geometry} and \citeA[\S 3.3]{melucci2015introduction}.

%\kirsty{One small change - a superposition *is* a pure state, so just changed it to superposition :)}
%\dom{Comments below stem from my previous use of the term "mixed states", which I admit I don't understand properly. I have for now changed the wording to "superpositions of pure states". Can discuss the use of the term mixed states further if we want to, just accept this version for now?}
%\kirsty{Are you sure about this? Back to my earlier points.. my understanding is that a pure state is essentially just a *single* state. That is, one that can be represented by a wave function. Mixed states are then just collections of pure states. Although it gets quite complex at this point because once you have a density matrix you can lose track of that collection, and get very very confused about what you have. D'Espagnat has a very good discussion on all this. Anyway - could this subspace not just be one for a single system? In which case it would be a pure state (things are much less complex when dealing with single systems - easier for me to make the call).} 

\subsubsection{Subspace Negation}

The negation of a subspace in quantum logic is not just its set complement, but its orthogonal complement: for example, for a plane in 3-dimensions, the orthogonal complement is the linear subspace normal to the plane. 
For example, in Figure \ref{fig:subspaces}, the orthogonal complement to the plane $A\vee B$ is the line $C$ that is 
perpendicular to both $A$ and $B$ and passes through the origin.
Negating a concept therefore involves projecting a vector onto this normal subspace, which can move it over large parts of the space, so in this sense quantum negation is a {\it non-local} operation.

\subsubsection{Subspace Conditionals Following Van Rijsbergen}

The material conditional $A \rightarrow B$ of classical logic is modelled by the inclusion $A \subseteq B$ in set theory,
and the direct quantum analogy of this is the subspace inclusion $A \leq B$, which means that $A$ is contained in $B$ {\it and} $A$ and $B$
are both subspaces. In terms of the corresponding projection operators, $A \leq B$ if and only if $\pi_B\pi_A = \pi_A$, which is to say that 
if we have already projected onto $A$, projecting onto $B$ does nothing.

In information retrieval, however, \citeA[Ch 5]{rijsbergen-geometry} is trying to describe the relationship between
a query and a relevant document, modelled by the document `implying' the query --- in the sense that the relevant
document contains {\it at least} the information in the query and more. Arguing that typically, a query is not {\it exactly} 
implied by any document, the notion of relevance-as-implication is qualified by the {\it Logical Uncertainty Principle} of \citeA{rijsbergen1986logic}:

\begin{quote}
    Given any two sentences $x$ and $y$; a measure of the uncertainty of $y\rightarrow x$ relative to a given data set,
    is determined by the minimal extent to which we have to add information to the data set, to establish the truth
    of $y \rightarrow x$.
\end{quote}
\citeA[Ch 5]{rijsbergen-geometry} argues that a conditional operator that meets this criterion is (the projection onto)
the subspace $A^\perp \vee(A \wedge B)$. This corresponds to the notion that an object supports the claim 
``birds fly'' if it is a bird that flies or if it is not a bird at all. (The fact that pigs don't fly doesn't provide
positive evidence that birds fly, but at least it does not refute the claim!) Van Rijsbergen's exploration is logical rather than 
empirical: the conditional is shown to satisfy logical compatibility properties explored by earlier authors, and in the 2004 work,
applications in information retrieval such as relevance feedback and clustering were proposed but not demonstrated.
Nonetheless, this work sets the stage for later developments, including the use of error terms in the classification
experiments of \citeA{bankova2019graded} and \citeA{lewis-2019-compositional} discussed below in Section \ref{sec:densitymatrices}.

%%%
\subsection{Application 1: Orthogonal Negation}

It turns out that these relatively simple observations can be applied directly to vector models for language. For example, suppose a corpus contains several uses of the word {\it suit} in the sense of {\it lawsuit}, and several in the sense of {\it garment}. It follows that the vector learned from training for {\it suit} will involve a superposition of these senses, so 
\begin{equation}
\label{eq:suit}
\mathit{suit} = \mathit{suit}_\mathit{lawsuit} + \mathit{suit}_\mathit{garment}.
\end{equation}

Now, suppose a user is searching for documents about {\it suit} in the sense of garment. An effective way to meet this need is to project the query vector from the term {\it suit} onto the subspace orthogonal to {\it lawsuit}. This technique was explored in detail with various term-retrieval measures  by \citeA{widdows-quantum}, which demonstrated that the immediate advantage of projection over Boolean negation in 
IR experiments was that it removed not only documents containing the unwanted term {\it lawsuit}, but other documents that use terms related to {\it lawsuit}.
\dom{This form of vector negation reduced the occurrence of synonyms and neighbours of the
negated terms by as much as 76\% compared
with standard Boolean methods  --- a practical outcome of the non-local nature of the negation operator.}
Its effectiveness in improving document retrieval was demonstrated by 
\citeA{Basile2011}\tc{, with relative increases of up to 10.8\% in Mean Average Precision (MAP) over a standard baseline obtained when using a re-ranking strategy involving rendering the vector average of 5 relevant documents orthogonal to that of a single irrelevant counterpart, and much greater increases when incorporating more relevance judgments (e.g. relative increase of 66.46\% with feedback from 40 relevant and 40 irrelevant documents).}
% Footnote disabled while Dominic and Trevor compared these - we think the `similar' results are contrasting a reduced `semantic space' with
% plain term-document vector space model.
%\footnote{Interestingly, the authors report similar performance improvements with a ``classical'' approach involving subtracting document vectors in a standard Vector Space Model. We note that these experiments employed a version of quantum negation in which a set of documents was represented by their sum, which raises the question of how a subspace-based alternative might perform on this task.}
The application of a projection operator to remove or emphasize aspects of a mixed semantic vector representation follows naturally from the fact that such vectors are generated as superpositions of vectors that represent occurrence in a (usually sense-specific) context. 
For a more thorough exposition of these points, see \citeA[Ch 7]{widdows-geometry} and \citeA{melucci2015introduction}. 

% TODO: Consider putting this back.
%These advantages are further reinforced by recalling the discussion of quantum probability in section~\ref{sec:probability}, and the manner in which a geometric approach is used to incorporate the notion of context.  In the context of a court of law the probability of a person interpreting the term \emph{suit} as an item of clothing can be expected to be decrease. Fig.~\ref{fig:contextualP}(b) depicts this state of affairs, in the $p$ basis we see that the probability of returning a $|1\rangle$ result is much smaller than it is in the $q$ basis. 
%On the other hand, if playing a game of cards then a person would be far more likely to interpret the term \emph{suit} in another manner entirely, which would be represented using an extra dimension in the basis. 
%This scenario becomes even more interesting when we consider the notion of compositionality that will be discussed in section \ref{sec:composition}.

Although negation is something of a niche operation in search engines, this example helps us to start motivating an answer to the question: ``What can quantum models do for artificial intelligence?'' In the case of modelling negation, quantum logic from the 1930's has provided a direct and computationally simple technique for performing logical operations in a vector model search engine. Building more thoroughly on the same foundation, \citeA{garg2019quantum} have recently used the rules of quantum logic (including negation)
to define objective functions for optimizing the representation of ontologies and relationships between concepts as vector space embeddings.
So from this point of view, quantum theory has motivated and guided these successes. 
However, it is partly a historical accident that the logic of projections and subspaces is called quantum logic at all: the meet, join, and orthogonal complement operations on a vector space were all discussed back in the 1860's by \citeA{grassmann-extension}, and the logical structure could easily have been explored independently of its application to quantum mechanics by Birkhoff and von Neumann in the 1930's. So perhaps the right question is not ``Can quantum logic be applied to IR?'' (it can), but ``Would it be better to refer to `vector logic' rather than `quantum logic' throughout the literature?'' 

\dom{The interpretation of products between vectors as subspaces is also a key part of {\it geometric algebra}, which can be used
a fundamental model for modelling several physical and computational operators \cite{dorst2010geometric}.}
One further complication here is that there are other logical systems with vectors, including those of \citeA{mizraji1994modalities}, \citeA{westphal2005logic}, and \citeA{clarke2012context}. The question of which vector logic works most effectively for information retrieval 
or other word embedding systems and applications is a research area that is likely to yield many interesting new results, but which has not to date attracted as much attention as it deserves.

%%
%\dom{Clean up these last few paragraphs (they were later in the Composition section but fit better here)}
%
%These models derive vector representations of terms from their distribution across a large text corpus, such that terms that occur in similar contexts will have similar vector representations. A typical example is the representation of terms occurring in a corpus of text  as the row vectors of a reduced-dimensional approximation of a term-by-document matrix, such as the matrix illustrated in Table \ref{term-doc-matrix-table}, in Latent Semantic Analysis (LSA) \cite{landauer-solution}. Similarities in meaning between terms can be estimated from their vector representations, most commonly using the cosine similarity metric. 
%
%Several authors have drawn attention to parallels between the mathematical structures used in such models, and those employed in quantum mechanics --- for example, the application of the Singular Value Decomposition in LSA and other semantic vector models corresponds to spectral decomposition \cite{aerts-semantic,bruza_quantum_2006}. 

%%%
\subsection{Application 2: Modeling Term Dependencies for Information Retrieval}

Another example of the utility of quantum models for information retrieval was provided by \citeA{sordoni_modeling_2013}, who investigated whether an extension of the standard unigram / bag-of-words language model to accommodate dependencies between terms in the context of frequently occurring multiword expressions such as {\it climate change}  
can improve retrieval performance. The issue of term dependencies is a longstanding problem in information retrieval. When constructing a probabilistic or geometric representation of a document, the question arises as to how to address the compositional nature of frequently occurring short phrases. A simple approach would involve treating the phrase {\it climate change} as a unit --- as though it were a single word. However, information about the occurrence of the component terms {\it climate} and {\it change} would then be lost. Furthermore, this approach provides no clarity about how best to weight such composite terms, which occur with less frequency than their unigram components. 

%\kirsty{suggest changing this example to a multiword expression that is not already being used all over the place here? Cute, but I think it is just going to confuse the reader. `Quantum theory' runs throughout this section so we need to change them all if we decide to do this.}  
%\trevor{done} 

To address these issues, Sordoni and colleagues developed a Quantum Language Model (QLM), a generalization of prior language modeling approaches that provides the means to model term dependencies without severing the connection between the probability of observing a multiword expression and the probabilities of observing its component terms. Here, we will provide a brief account of the model and the accompanying empirical results. We refer the interested reader to \citeA{sordoni_modeling_2013} for further implementation details such as the selection of phrases to model, smoothing and the estimation of model parameters.

%Language modeling approaches to information retrieval model units of text as samples from a probability distribution over terms in a vocabulary. The parameters of this distribution indicate the probability of observing each term in the vocabulary, and are estimated from the terms observed in a particular unit of text. Consequently, queries and documents can be compared using these estimated parameters --- for example, by estimating the probability of observing a particular query given a set of parameters learned from a document, with smoothing to account for unobserved terms \cite{ponte_croft_1998}; or by measuring the divergence between the parameters learned from a query and those learned from a document \cite{lafferty2001document}. 

Language modeling approaches to information retrieval can be represented geometrically by mapping each term in the vocabulary to an independent basis vector (familiar to readers with a machine learning background as a `one hot encoding' \cite[Ch 2]{geron2019hands}). Projection operators for each term can then be defined  to recover the probability of observing a particular term in the context of a set of language model parameters learned from a document. Projection operators for single words are diagonal matrices with a single non-zero value corresponding to the basis vector of the term concerned. 

With the QLM, multiword expressions are represented as a weighted superposition of the basis vectors corresponding to their component terms:  
\[
    \overrightarrow{climate\_change} = \alpha \times \overrightarrow{climate} + \beta \times \overrightarrow{change}
\]
which are normalized to length one. The projections onto these vectors lead to off-diagonal entries in the corresponding matrices, so the
term dependency cannot be represented as just a weighted sum of the `one-hot' diagonal elements. 
The coefficients chosen for this weighting encode the extent to which observation of the composite phrase indicates the presence of each component term. This is intuitively appealing, and provides the means to heuristically weight the relative importance of the terms within a phrase such that, for example, documents containing the term `climate' in isolation are more likely to be retrieved in response to the query `climate change' than those containing the term `change' alone. Given a unit of text, model parameters are learned using an approximation algorithm that attempts to find the parameters that maximize the probability of the observed document or query. Relevance ranking is performed on the basis of the divergence between parameters learned from queries and documents. Evaluation of the QLM was conducted using selection of 450 queries drawn from across four information retrieval evaluation sets. The best performance for each of the four sets was obtained by a variant of the QLM, outperforming a unigram language model baseline, with statistically significant improvements over then state-of-the-art approaches using Markov Random Fields (MRF) \cite{metzler2005markov} for the two larger web-based sets \tc{with relative increases in MAP for the best-performing QLM as compared with the best-performing MRF of 5.5\% and 5.2\%}. 

In summary, using quantum theoretic ideas directly in the traditional vector model for information retrieval has provided several concrete opportunities, 
including an account of conditionals and implications, negation and disjunction, and term dependencies, some of which have improved performance in retrieval experiments.

%Bose-Einstein statistics \cite{amati.vanrijsbergen:probabilistic}.

%%%
\section{Categories, Hypernyms and Implication: More Advanced Structures in Vector Spaces}
\label{sec:categories}

Having presented examples of how vector spaces can provide a rich set of tools for problems relevant to AI and ML, it is now time to move onto more advanced vector space structures --- what insights do they provide? Representing categories and hypernym relationships will help us to motivate the use of 
structures in vector spaces that are more complex than the individual points referenced by vectors that we explored in the previous section.

\subsection{Negation and `Parts' of Vectors}

The example above that uses negation to remove unwanted meanings of ambiguous words carries an important conceptual lesson:
 a single vector can represent multiple meanings in a high dimensional space, and specific ones can sometimes be recovered explicitly
using a method such as orthogonal projection to a subspace. 

This runs counter to a common belief expressed in the literature, that one point can represent only one thing. As an example of this line of thinking, consider  \citeA{neelakantan2014efficient} who start their widely cited paper with this statement:

\begin{quote}
There is rising interest in vector-space
word embeddings and their use in NLP,
especially given recent methods for their
fast estimation at very large scale. Nearly
all this work, however, assumes a single vector per word type --- ignoring polysemy and thus jeopardizing their usefulness for downstream tasks.
\end{quote}

\noindent
%The paper itself creates word vectors by clustering contexts of word occurrences (as pioneered by \citeA{schutze-automatic}) and builds a representation for each sense. 
The effectiveness of vector negation for uncovering different word senses in a single vector
demonstrates a flaw in the claim that a single vector per word ignores polysemy, 
even though the clustering results described by \citeA{neelakantan2014efficient} are still valuable. Mathematically,
this claim would be  equivalent to the claim that a single vector per particle cannot represent a superposition of pure states, a statement
which quantum mechanics has shown to be flawed in the quantum realm. Nonetheless, other authors have echoed this claim, 
including for example \citeA{camacho2018word}:

\begin{quote}
The prevailing objective of representing each word type as a single point in the semantic
space has a major limitation: it ignores the fact that words can have multiple meanings
and conflates all these meanings into a single representation.
\end{quote}

The assumption that
a vector is equivalent to a point and therefore cannot represent something with internal structure or ingredients goes back to the
beginning: ``A point is that which has no parts'' is Book I, Definition I of Euclid's {\it Elements}.  
However, even in Euclidean space, a vector can also be thought of as an arrow or straight path from the origin to a particular point. This straight path
certainly can have `parts'; in particular, each coordinate represents a length along a particular axis (in Fourier analysis, a basic harmonic function; in quantum mechanics, a basis state, etc.). 

The notion that vectors have parts is much more obvious in quantum theory, because as introduced in Section \ref{sec:qm-intro}, the vectors in quantum mechanics
are functions: and of course, a function can have many different parts. Fourier analysis itself is a particularly canonical way of breaking a function down 
into a sum of different parts. From this point of view, the mistaken belief that a vector can only represent one thing would
be unlikely ever to have arisen. 
Vectors can represent many types of things in many ways, and high-dimensional vectors in particular can accurately represent many
ingredients in such a way that the ingredients can be clearly recognized and sometimes recovered from the combined representation.
The mathematical foundations behind this claim are analyzed more thoroughly by \citeA{kanerva_hyperdimensional_2009} and \citeA{widdows2015reasoning}. 

Despite its mistaken foundations, work motivated by this understanding of vector models has provided useful results. The perspective of \citeA{neelakantan2014efficient} that we criticised above still created word vectors by clustering contexts of word occurrences (as pioneered by \citeA{schutze-automatic}) and built a representation for each sense. Similarly, the survey of disambiguation techniques by 
\citeA{camacho2018word}, 
and the smoothed Gaussian representations of \citeA{vilnis2015word} have both provided valuable contributions to the field.  Still,
the stated mathematical motivation for such work is not well-founded, and a quantum perspective on the mathematics naturally avoids this mistake, and provides a richer formulation that can potentially be leveraged to achieve novel results.

\subsection{Subspaces, Disjunctions, and Generalization}
\label{sec:subspaces}

Even though a single vector can represent several senses of a polysemous word, 
it is important to realise that individual vectors cannot mathematically represent everything that can be expressed in a vector space. 
The case of linear subspaces, described in Section \ref{sec:quantum-ir}, is an obvious example. A $k$-dimensional subspace
in an $n$-dimensional space typically requires at least $\min(k, n-k)$ vectors to describe it. The $n-k$ representation itself provides an interesting insight: it is 
useful when $k > \frac{n}{2}$, which makes it more convenient to express a subspace in terms of a normal projection to the subspace. This technique is most commonly encountered in 3 dimensions
where the normal to a plane is just a line (for example see again Figure \ref{fig:subspaces}, where the normal
$C$ could also be used to {\it define} the plane $A\vee B$, because $p\in A\vee B$ if and only if $p\cdot C = 0$).

The link between disjunctions and categories or other `natural kinds'  is that a more general
category like {\it mammal} arises as a disjunction of examples like {\it cat}, {\it dog}, {\it mouse}, {\it elephant} 
\cite[Ch 3]{rijsbergen-geometry}. 
In lattice theory, a disjunction is a `least common ancestor',  
characterized by being the most specific element available that subsumes (implies) all of the constituents \cite[Ch 8]{widdows-geometry}.
Since subspaces are the natural representation of disjunctions in quantum logic, it is tempting to assume they are a
good representation for categories of word vectors. This intuition has worked in some cases: for example, 
a combined representation built using
quantum subspace disjunction with limited numbers of inputs has been shown to perform well at the task of geometrically-mediated analogical inference, generally recovering more therapeutic relationships between drugs and types of cancer than comparable superposition-based approaches \cite{cohen-many-paths} and in information retrieval, subspaces have been shown to be particularly effective for modelling {\it negated} disjunctions --- that is, removing many unwanted 
areas of meaning from a query vector \cite{widdows-negation}.
However, a mathematical problem arises with positive disjunctions
of many inputs: linear sums of many similar vectors tend to extrapolate and overgeneralize \cite{bruza2009quantum}.
Specifically, a disjunction of $k$ non-degenerate vectors {\it always} leads to a $k$-dimensional subspace, 
however similar or different the initial vectors are. For example, back in Figure \ref{fig:subspaces}, $A$ and $B$ would generate the
exactly same plane $A\vee B$ whether they are close together or far apart, so long as they are not identical.
This also means that  in an $n$ dimensional space, $n$ very slightly different examples 
would nearly always generate the whole space. So while subspaces have been shown to work well for some forms of semantic
generalization, we should expect this to be an incomplete and eventually over-general model.

\subsection{Density Matrices, Positive Operators, and Hyponymy}
\label{sec:densitymatrices}

Quantum mechanics is already familiar with the problem that individual vectors and even subspaces cannot 
represent everything we encounter in practice. A single particle in a superposition of pure states can be
represented as a single vector, but an ensemble of particles cannot. Instead, ensembles are modelled
using a {\it density matrix} or {\it density operator}, which takes the form

\[ \rho = \sum_j P_j \ket{\Psi_j} \bra{\Psi_j}. \]

\noindent
In quantum mechanics, a system of particles that can be expressed as a density matrix but not a state vector is
called a {\it mixed state}.
The motivation for generalizing from states to density matrices is discussed in detail by  \citeA[Ch 6]{isham:quantum}, and 
\citeA[Ch 6]{rijsbergen-geometry}. Mathematically, the formalism is general enough to represent all the relevant
probability distributions, which an individual state vector $\ket{\Psi}$ with corresponding density operator
$\rho = \ket{\Psi} \bra{\Psi}$ cannot.

Density matrices are positive operators, in the sense that $\langle \Psi | \rho | \Psi \rangle \ge 0$ for all vectors $\Psi$.
This can be used to induce an ordering on density operators, defining $A \sqsubseteq B \Longleftrightarrow B - A$ is positive.
This is called the L\"owner ordering, and when the operators are restricted to projection onto subspaces, the L\"owner ordering
becomes equivalent to the quantum logic introduced in Section \ref{sec:subspacesQL}.
The L\"owner ordering is used by \citeA{bankova2019graded} and \citeA{lewis-2019-compositional} to represent
graded hyponymy (where, for example, {\it dog} may be said to be a strict hyponym of {\it mammal} but a graded hyponym of {\it pet},
because not all dogs are pets). In cases where $A$ and $B$ are incomparable in this ordering, the
systems proposed in these papers
work by finding positive operators $D$ and $E$ (an error term) such that $A + D = B + E$.
This avoids the problem with subspaces whereby any part of $A$ that is not part of $B$ adds whole new dimensions: instead,
if $A$ is nearly subsumed by $B$, the error term $E$ will be correspondingly small.
The smaller $E$, the more strict the hyponymic relationship. In further work, \citeA{lewis2020towards} extends this system
to produce a graded form of negation, which can be applied to more situations than the orthogonal negation of \citeA{widdows-negation}.

This example is critical, and illustrates the main point of this section: while vectors are powerful representations and can represent
many ingredients at once, they cannot represent (for example) all the probability distributions necessary for quantum 
mechanics. Exploring the ways these shortcomings have been addressed within quantum theory suggests potentially fruitful research avenues for those using vectors to model concepts in language.

%%%
\section{Products of Vectors and Semantic Composition}
\label{sec:composition}

The summary of categorization and negation in the previous section highlights  one of the key places 
where quantum techniques are useful in AI. They provide us with  a range of operators for exploring, 
manipulating, and generating representations of concepts using semantic vector models. Some of these are familiar, some are novel, and all open up potential new avenues for research in AI. 
In this section we draw attention to the problem of semantic composition. 
There are many ways to compose vectors --- so far in this paper we have discussed the vector sum, and various operators related to subspaces
and projections.
More complicated structures are available, and semantic composition is often represented in vector models using the tensor product and its offshoots. 

%%%
\subsection{Tensor Products and Entanglement}

The tensor product is one of the most significant methods used for composing vectors.
After exploring addition and subtraction, the next product structure on vectors defined by 
\citeA[Ch 2]{grassmann-extension}
is the forerunner of today's tensor product in coordinate form: if $a = \sum a_r e_r$ and $b = \sum b_s e_s$,
then their product $[ab]$ is defined as $\sum a_r b_s [e_r e_s]$.
By varying the rules for interpreting and identifying the basic terms $[e_r e_s]$, Grassmann 
showed that such product structures can be used to represent 
the inner (scalar) product and combinatorial (exterior) product.
Note that the order matters here: unlike the sum of two vectors or the product of real or complex numbers, this 
product between two vectors is not commutative. This makes it more complicated, but also opens opportunities.
For example, generating a document vector as a weighted sum of term-vectors is surprisingly effective
for information retrieval, but a commutative sum that fails to take word-order into account is unsuitable for building a number of important applications (e.g.  a conversational agent or any other dialogue system).

We have already noted that vectors can represent operators as well as states --- starting with the (co)vectors
in the dual space $V^*$ that act as linear maps from $V$ to its ground field (the {\it ground field} being
the number system used for coordinates, most commonly the real or complex numbers).
Given this, Grassmann's product operator
can be used to represent a bilinear map from $V\times V$ to the ground field --- for $\alpha, \beta \in V^*$ and $u, v \in V$,
we define $[\alpha \beta] (u\times v) = \alpha(u)\beta(v)$, and it is easy to see that this map is linear in all of its 
arguments. In Dirac notation, a product $\bra{\phi_i \phi_2}$ 
of the bra vectors $\bra{\phi_1}$ and $\bra{\phi_2}$ would map the product of two ket vectors $\ket{\psi_i\psi_2}$
to $\braket{\phi_1|\psi_1}\braket{\phi_2|\psi_2}$.

Today the space of all possible linear combinations of products of basis vectors of two Hilbert spaces is called their 
{\it tensor product}, written using the symbol $\otimes$, so that the tensor product of vector spaces 
$U$ and $V$ is written $U\otimes V$.
It is defined more formally as an equivalence class of mappings whereby (for example) if $E(A, B) \rightarrow U$ is a bilinear map
from (the Cartesian product of) $A$ and $B$ to $U$, then this corresponds to a unique linear mapping from $A\otimes B$ to $U$,
and it turns out that all bilinear maps from $A\times B$ to $U$ can be represented in this manner. Linear mappings from one space to another
have a similar correspondence: the space $L(A, B)$ of linear transformations from $A$ to $B$ is naturally isomorphic to $A^*\otimes B$
\cite[Ch. 16]{lang-algebra}. 

In a given coordinate basis, the tensor product of two vectors can be represented as a matrix with the same coordinates as their
{\it outer product}. Just as the inner product or scalar product of two column vectors $u$, $v$ can be written as $u^T v$, their outer
product can be written $u v^T$. An example with vectors in $\mathbb{R}^3$ is the following:

\[
u = \left( \begin{array}{c} 1 \\ 0 \\ -2 \end{array} \right) \qquad
v = \left( \begin{array}{c} 2 \\ -1 \\ 3 \end{array} \right) \qquad
u^T v = 2 + 0 - 6 = -4 \qquad
u v^T = \left( \begin{array}{ccc} 2 & -1 & 3 \\ 0 & 0 & 0 \\ -4 & 2 & -6 \end{array} \right)
\]

\noindent
While this is intuitive and familiar, the mental identification of tensors with matrices does lead to gaps. 
For example, the notation $u^T v$ for the scalar product in coordinate matrix form encourages us to think of 
row vectors as dual to column vectors, but matrix multiplication alone does not enable us to map a $(n, 1)$ column vector
to a $(1, n)$ column vector. 
(Remember that matrices of size $(p, q)$ can only be multiplied on the right by matrices of size $(q, r$), resulting 
in a matrix of size $(p, r)$: so a matrix of size $(n, 1)$ can only be multiplied on the right 
to give a matrix of size $(1, n)$ if $n=1$, with a similar argument holding for left-multiplication.)
So if we were to identify tensors with such matrix representations, we would be tricked into believing
that tensors cannot be used to map vectors to covectors, which is not the case. 

More generally, tensors form an algebra of multilinear maps. If $V$ is a vector space and $V^*$ is its dual, then
the tensor space $V\otimes \ldots \otimes V \otimes V^* \otimes \ldots \otimes V^*$ consists of multilinear functions from
$V^* \times \ldots \times V^* \times V \times \ldots \times V$ to the ground field. (The $*$-symbols are deliberately switched between the first
and the second expression: the elements of $V^*$ act on elements of $V$, and vice versa.) 
Such a product with $p$ copies of $V$ and $q$ copies of $V^*$ is called a {\it tensor of type} $(p, q)$,
and the number $p+q$ is sometimes called the {\it rank} or the {\it arity} of the tensor.
For example, the density matrices of Section \ref{sec:densitymatrices}, being the product of a bra and a ket vector,
are tensors of type (1, 1) and rank 2. 

Tensor algebra has been used extensively in many fields before computer science. For example, it has been in use in differential geometry since the early 1900's, partly because so
many geometric concepts can be expressed easily using tensors, including vector fields, differential forms, 
metrics, volume integrals, complex structures and Hamiltonian dynamics. 
Because of this, some of the most readable and thorough introductions to tensor algebra are from standard 
texts in differential geometry (for example, \citeA[Ch 5, 6]{willmore1959introduction}, \citeA[Ch 2]{bishop1968tensor}). 
Continuum mechanics, which models the behavior of materials and liquids as a continuous mass, also makes extensive use of tensor algebra to extend Hooke's law to high dimensions in the modelling of liquids, elasticity of materials, and other important phenomena (via the stress-strain tensor --- see \citeA{spencer2004continuum}, which includes a good introduction
to matrix and tensor algebra for mechanics). 
Finally, as we have seen,
in quantum mechanics the tensor product naturally arises when we move towards representing composite systems. 
More recently, in machine learning,
tensors have made a sustained contribution  
through the introduction of systems such as TensorFlow for training neural networks \cite[Ch 12]{geron2019hands}.
Tensor products are also used to describe the very general notion of `processes happening simultaneously' in
physics, sometimes generalizing the notion of tensor product to mathematical settings beyond linear algebra \cite{coecke2017picturing,heunen2019categories}.

In summary, tensors have become an invaluable part of practical mathematics in many fields. However, this very multidisciplinarity 
can result in confusion around terminology.   For example, a matrix can be used to represent a rank-2 tensor,
but then the definition of the term {\it rank} is different from the traditional definition in linear algebra, where the rank of
a matrix is the number of linearly independent rows and columns (a key theorem in linear algebra guarantees that the row rank and
the column rank are the same, so it makes sense just to talk about the rank of a matrix \cite[Ch 5]{janich-linear}). 
This notion of rank also generalizes to tensor algebra, and the rank of a tensor is
also used to mean the number of linearly-independent generators for the tensor. In some fields (including machine learning),
the difference between vector spaces and their duals does not yet play nearly as important a role as it does in differential geometry,
and so this difference is often glossed over, and instead of talking about tensors of type $(p, q)$, only the tensor
rank $r = p+q$ is discussed. In particular, it is often said that the rank-2 tensors are equivalent to matrices,
which we have seen is an algebraically incomplete representation (in the example above, a linear mapping from vectors to covectors can be 
represented as a rank-2 tensor of type $(1, 1)$, but no matrix can be found that maps column vectors to row vectors using multiplication).
This teaching may change gradually, because used carefully, the distinction between
vectors and covectors can be a benefit in machine learning as well. For example, \citeA{turney2012domain} 
used precisely this approach to make a useful distinction between domain and functional similarities in distributional semantics.

\subsection{Tensor Products and Entanglement in Quantum Mechanics}
\label{sec:tensor_qm}

In quantum mechanics, superposition becomes incomplete for describing the state of systems 
as soon as we consider composites involving more than one particle. 
Consider, for example, particles with wavefunctions $\ket{\phi}$ and $\ket{\psi}$,
each represented as a superposition $\sum_1^m a_i \ket{\phi_i}$ and $\sum_1^n b_j \ket{\psi_j}$, where $\ket{\phi_i}$ and $\ket{\psi_j}$ are
the eigenstates for some observables $A$ and $B$. In cases where the particles can be observed simultaneously, the result
of measuring $A$ and $B$ together can be any combination of $\ket{\phi_i}, \ket{\psi_j}$, and there are $mn$ such possibilities.
Therefore the combined system of $\ket{\phi_i}$ and  $\ket{\psi_j}$ can consist of any linear combination
of these eigenstates, so it follows that the Hilbert space of possible states for this system has $mn$ dimensions.
This deceptively natural conclusion should be contrasted with the classical situation, where the state space for 
the combined system would be the Cartesian product of the state spaces for the individual systems, having dimension $m + n$.
For a step-by-step example with spin states of 2- and 3-dimensional particles, see \citeA[\S 8.4]{isham:quantum}.
The combined state where measuring $A$ results in $\ket{\phi_i}$ and $B$ results in $\ket{\psi_j}$
is written as $\ket{\phi_i \psi_j}$, and the state that represents the combination
of the superpositions $\sum a_i \ket{\phi_i}$ and $\sum b_j \ket{\psi_j}$ is 
$\sum a_a b_j \ket{\phi_i \psi_j}$, which is exactly the same as Grassmann's definition of the product of two vectors above,
just rewritten in Dirac notation.

The mathematical behavior of tensor products is directly responsible for
the famous phenomenon of {\it quantum entanglement}.
Entanglement is perhaps easiest to introduce 
via its opposite phenomenon; separability. A composite state is considered separable if it is possible to write it as a tensor product of two distinct
states (i.e. as $\ket{\phi}\otimes\ket{\psi}$ 
for some $\ket{\phi}$ and $\ket{\psi}$), in which case it is generally considered to be non-interacting.
In contrast, an entangled state cannot be decomposed in this manner.

How does this come about? Mathematically, the answer is most obvious if we consider the dimensions involved. If $U$ is of dimension
$m$ and $V$ of dimension $n$, their tensor product $U\otimes V$ has dimension $mn$, which of course is typically greater than $m + n$.
Immediately it follows that not every element of $U\otimes V$ can be written as some individual product $u\otimes v$, because there are only
$m + n$ degrees of freedom for choosing the $u$ and $v$. In order to generate any element of $U\otimes V$, we may need to superpose
many different individual products of the form $u\otimes v$, just as in the example system above where the product state must be written
as a linear combination of states like $\ket{\phi_i} \otimes \ket{\psi_j}$. 
A typical way of explaining the difference algebraically is that $u \otimes v$ is not a standard {\it element } of $U\otimes V$, but it is
a standard {\it generator} or basis vector for $U\otimes V$. In matrix algebra, this corresponds to the result that any rank-1 matrix
(in the sense of matrix rank above!) can be written as the outer product of a single row vector and a single column vector.

With superposition and quantum disjunction, we saw earlier that a quantum system might be represented
as the sum of its parts without being identical with or contained in any of those parts. The tensor products allows
for even richer combinations where the product is not contained in any of the ingredients, and cannot even be
broken down into a combination of one simple ingredient from each part. A vector $u$ in the subspace $A + B$ can at least 
be represented as the sum $a + b$ for some $a \in A, b \in B$, and this decomposition is easily obtained using projection operators.
On the other hand, a tensor $u \in A\otimes B$ cannot even be decomposed like this: at its most general, it must be written as
a linear combination of the form $u = \sum a_i \otimes b_i$ (and the minimum possible $i$ is the tensor generalization of the rank of a matrix).
While quantum logic can be described as non-distributive and non-local, tensor products and entanglement
demonstrate {\it non-separability} \cite[\S 4.1.1]{coecke2017picturing}.

%\dom{I recommend using the matrix form above rather than the Dirac notation that I've commented out below. Thoughts?}\kirsty{yes - I think  it would be good to be consistent, and better to use matrix form as that will be understood by AI people more easily}\kirsty{Actually - on reading the whole paper through I might have just changed my mind. We have been using kets to represent states right from the start of this paper.. should be ok?}
%Thus, a composite state $|\psi\rangle_A\otimes|\psi\rangle_B$ arises in the Hilbert space describing the composite system $H_A\otimes H_B$. More specifically, if we can write each state in terms of a general basis
%\begin{align}
%|\psi\rangle_A = \sum_i c_i^A |i\rangle_A \mbox{ where } |i\rangle_A \in H_A \mbox{ and }|\psi\rangle_B = \sum_j c_j^B |j\rangle_B \mbox{ where } |j\rangle_B \in H_B 
%\end{align}
%where there exist vectors $c_i^A, c_j^B$ so that $c_ij=c_i^Ac_j^B$ for the general state 
%\begin{align}
%|\psi\rangle_A\otimes|\psi\rangle_B &= \sum_i c_i^A |i\rangle_A \otimes \sum_j c_j^B |j\rangle_B \\
%|\psi\rangle_{AB}& = \sum_{i,j} c_{ij} |i\rangle_A\otimes |j\rangle_B
%\end{align}
%then that state can be defined as compositional. If no such $c_i^A, c_j^B$ exist, then the state is said to be entangled.

%In quantum mechanics, a product state $\ket{\phi\psi}$ that can be written directly as a tensor product $\ket{\phi}\otimes\ket{\psi}$ for some $\ket{\phi}$ and $\ket{\psi}$ is called {\it separable}. A product that cannot be written in this way is called {\it entangled}. 

The concept of entangled quantum systems lies at the center of some of the most interpretationally fraught aspects of
quantum mechanics \cite{isham:quantum,greenstein.zajonc:quantum}. For example, if a system of two particles is  entangled, but then one of its constituent particles
is measured and thus observed to be in an eigenstate, what does this mean for the combined system and for the other particle? 
%This question led to the famous Einstein-Podolsky-Rosen thought experiment, which was designed to demonstrate the nonsensical nature of quantum mechanics itself. 
While 
physical implications of entanglement are conceptually challenging, the 
concept is well-enshrined in modern physics: creating and observing entanglement between qubits (quantum bits) is one of the key 
necessary and challenging processes in quantum
computing, a point that we will return to in Section~\ref{sec_implementation}. First however, it is worth highlighting some of the ways in which this theoretical phenomenon has been exploited in language models in AI.  
\subsection{Explicit Composition with Vectors and Tensors in AI}

The use of tensor products in AI is often traced to the work of \citeA{smolensky1990tensor}, and its uses
have become much more widespread more recently. During the later decades of the 20th century, vector models were often
described as ``bag of words'' models because of the commutative nature of the 
vector sum operation they applied. This choice of operator meant that  the same representation could be used for a collection of words in 
a document, irrespective of the order in which the words appear. 
By contrast, discrete logical models used in formal semantics have for many decades been quite explicit about
the ways words should be combined, but were often notably silent about what those words mean in themselves
(see \citeA{widdows-products,baroni2014frege} for surveys of this methodological difference between traditions).
This history of two modelling frameworks led to an  unnecessary and unfortunate gap: there are many interesting product operations such as the tensor product between vectors
that are well-established in linear algebra, but for years there were relatively little awareness of these
alternatives in language research. Since the early 2000's, this situation has changed markedly.

In a pioneering case for AI, tensor products were introduced at least as early as the work of \citeA{smolensky1990tensor}, where the tensor
product was used to bind a variable to its value, and the term
`binding' for products of vectors has been used in various language and cognitive models involving vectors since the work of  \citeA{kanerva_hyperdimensional_2009}.
In the first decade of the current century, the use of tensor algebra for combining word
vectors was explored using a quantum formalism by \citeA{aerts-semantic}, whereby a sentence 
$(w_1, \dots, w_n)$ would be represented as the tensor product $w_1 \otimes  \ldots \otimes w_n$.
Taking this in explicitly linguistic directions, product operations on vectors were introduced to model grammatical and semantic composition 
\cite{clark2007combining,widdows-products}. Particularly interesting examples included the work of 
\citeA{mitchell2010composition} on noun composition, and \citeA{baroni-nouns} on adjective-noun combinations.
\citeA{turney2012domain} used a combination of two vector spaces (a large-window space 
capturing domain similarities, and a narrow-window function space capturing functional similarities) to combine relational similarity with two-word compositional behavior.

% More recent work has applied vector operators (other than superposition) both to compose representations of
% larger units of text, such as sentences, from semantic vector representations of terms \cite{heunen_quantum_2013}, 
% and to encode additional information into semantic vector representations 
% \cite{jones-holographic,devine-semantic,symonds:tensor}.

Inspired by Smolensky's work on binding operators, the model of \citeA{clark2007combining} used the tensor product as a means to generate vector representations of phrases, noting that one desirable property of the tensor product in this context is that it does not commute, so the product $\mathit{Djokovic} \otimes \mathit{beat} \otimes \mathit{Murray}$ is not equal to the product $\mathit{Murray} \otimes \mathit{beat} \otimes \mathit{Djokovic}$. 
Another useful mathematical property is that the inner product of two tensors is equal to the product of the inner products of their related constituents, e.g. $(\mathit{beat} \otimes \mathit{Murray}) \cdot (\mathit{defeated} \otimes\ \mathit{Andy})  = (\mathit{beat} \cdot\mathit{defeated}) \times (\mathit{Andy} \cdot \mathit{Murray})$. This means that the similarity between a pair of phrase tensors can be calculated without the need to explicitly represent them (which would require $n^2$ dimensions for $n$ dimensional vectors), and, with normalized vectors, this is the product of the cosine similarities between the component vectors for terms that occupy the identical position in the sentence structure. However, the grammatical structure of related sentences may not be identical, and as such their composite representations cannot be easily compared with this method. 

A solution to overcome this limitation of the tensor product approach was proposed by 
\citeA{coecke2010mathematical}. They made use of category theory \cite{lawvere2009conceptual} to demonstrate that vector spaces and certain types of grammar (exemplified in this work by pregroup grammars \cite{lambek2001type}) fall under the same category type, compact closed categories (see Appendix \ref{sec:category_theory}.
This realization provided a means to map between the grammatical and the vector representation, 
and thus demonstrated that the rules for grammatical composition can be applied to a compositional model in vector space.  
The mathematical roots of this work are directly related to quantum mechanics due to the work of \citeA{abramsky2009categorical}, which developed {\it Categorical Quantum Mechanics}, 
an axiomatic presentation of quantum theory using compact closed categories, which makes the relationship with the grammatical work of Lambek particularly clear. (An online appendix is provided for readers unfamiliar with category theory.)
This point will be revisited in Section \ref{sec_implementation} because it is especially relevant to implementation on quantum computers. 
Since its introduction by \citeA{coecke2010mathematical}, the Distributional Compositional Categorical model has become known as DisCoCat.

\citeA{grefenstette-experimental} provided an implementation and evaluation of this approach, deriving word representations from the British National Corpus. 
To do so, different rules of composition were assigned to words with different grammatical types in accordance with the categorical type of their pregroup, with `atomic' types such as nouns represented by  distributional term-by-context-term vectors, and `adjoint' types generated compositionally. For example, verb representations were generated as the sum of the tensor products of the noun vectors corresponding to their subject and object across every occurrence in the corpus (intransitive verbs and adjectives can also be accommodated). Vector representations of sentences can then be compared: for example, the vector representation of the sentence ``Djokovic beat Murray'' is composed by pointwise multiplication of the tensor representing the verb ``beat'' and the tensor composed from the vectors for ``Djokovic'' and ``Murray'': $(\mathit{Djokovic} \otimes \mathit{Murray}) \odot \mathit{beat}$. The model was evaluated for its ability to estimate the similarities between short phrases, and correlated with human judgment of similarity as well as, and better than, the best available models on datasets %concerning adjective-noun and subject-verb-object relationships respectively. 
concerning verb-noun and subject-verb-object relationships. \tc{The authors report Spearman correlations of 0.17 and 0.21 for the verb-noun and subject-verb-object sets respectively with the compositional model, as compared with 0.17 for both sets with the multiplicative model of  \citeA{mitchell2008vector}.}

In methodologically similar work, words such as verbs and adjectives that take arguments have been represented as tensor products in matrix form. For example, \citeA{baroni-nouns} used this approach to model the action of adjectives upon nouns, and
\citeA{socher2012matrix} took it to the logical destination of representing
each internal node in a parse tree as a matrix operator acting upon its
input arguments. This area has become known as Compositional Distributional Semantics, summarized by \citeA{baroni2014frege}, and work in this area has continued, an example being the work of \citeA{sadrzadeh2018sentence} on sentence entailment in this framework,
\dom{where an F1-score of 0.86 is reported on recognizing semantic entailments like ``robin flies $\models$ bird moves''.} 

As with vector logics, the tensor product as used in quantum mechanics is only one of the composition operations that can be used with vectors,
albeit an important one. One implication of using the tensor product is the inevitable explosion in dimensionality that arises: a tensor product 
takes two $n$-dimensional vectors and makes an $n^2$-dimensional tensor, which is a problem for scalability on classical hardware, and creates products that are not obviously comparable with their inputs. Various algebraic structures, sometimes known as Vector Symbolic Architectures \cite{levy2008vector}, 
have been used to address this problem. They make use of  circular correlation and circular convolution, 
to roll the tensor product coordinates back into coordinates in
the initial space. The use of these structures  for reasoning in continuous models
has been described by \citeA{widdows2015reasoning}.

Entanglement has also been used in compositional modelling for language concepts,   
motivated partly by the study of entanglement for word-association in cognitive models \cite{bruza2008entangling,aerts2011quantum}.
For example, \citeA{kartsaklis2014study} explore the entangled
representation of transitive verbs using tensors in the categorical framework of \citeA{coecke2010mathematical}.
In an explicit use of entangled superpositions for reasoning and inference,
predication-based semantic analysis was developed by \citeA{cohen-many-paths}, and makes use of vector binding operations to represent a concept $A$ occurring in a relation of the form $R(A, B)$ as the sum of the products $r\otimes b$ for each relation in which it occurs. 
(For example, with the relational triple `aspirin TREATS headache', the concept vector for {\it aspirin} 
gets incremented with the bound product vector $\mathit{TREATS}\otimes \mathit{headache}$.)
In this framework, concepts become represented as superpositions of products of pairs, and these  cannot
be expressed as any product of single ingredients $r$ and $b$. This uses some of the same mathematics that in quantum mechanics leads to entanglement, including the use of complex vectors as a ground field in some implementations \cite{cohen-many-paths,cohen2015embedding_probabilities}. (In an alternative approach to modelling relations using complex numbers, the work of \citeA{garg2019quantum} explicitly uses the complexified representation $A + iB$ to form pairs, and axioms of quantum logic to formulate the compositional representation of relations.)

From the point of view of language and relationships, the notion of a general relation being derived from several varied
examples makes intuitive sense. 
Each of the pairs ({\it Leto}, {\it Artemis}\,), ({\it Henry VIII}, {\it Elizabeth I}\,), ({\it Lord Byron}, {\it Ada Lovelace}\,), ({\it Darth Vader}, {\it Princess Leia}\,), ({\it Debbie Reynolds}, {\it Carrie Fisher}\,),
is an example of a parent-child relationship, and all of these could be combined into a thorough and very varied notion of parenthood.
Because of this variety, it would be surprising if there was any one `prototypical parent' and `prototypical child' 
that combines to make the relation of `prototypical parenthood'.
Thankfully, the mathematical structures used for tensor products that give rise to entanglement enable us to represent
a relation that is a rich combination of different example pairs,
even for relations where no single pair of prototypical ingredients exists \cite[\S 5]{widdows2015reasoning}.

\subsection{Implicit Composition in Deep Neural Networks}

Something the methods described in the previous section have in common is that
they encode some {\it explicit} syntactic structure: a role / value binding,
or a relationship in a grammatical parse tree. This raises the question: can
models go beyond the bag-of-words drawbacks and encode more order-dependent
language structures without using this traditional syntactic machinery? 
A recent and comprehensive survey of this topic is provided by \citeA{hupkes2020compositionality}.

During the same years that compositional distributional
semantics has been developed, neural networks have made great strides in 
artificial intelligence, particular the use of networks with several intervening layers, 
hence the term `deep learning' \cite{lecun2015deep}. 
In some cases, work on deep learning and compositional semantics has been
explicitly combined: for example, \citeA{socher2012matrix} describe Recursive 
Neural Networks used for training. These should not be confused with the
the simpler and more standard Recurrent Neural Networks (RNNs), in which
the output of a single neuron depends on its inputs and its previous state. RNNs have been used for many sequence-modelling
tasks, as have their more sophisticated cousins, LSTMs (Long Short-Term Memory cells, where prior state may be stored for 
longer and updated based on learned importance) \cite[Ch. 15]{geron2019hands}.

Later, attention-based networks have been introduced, where the attention mechanism is designed to learn when pairs of inputs depend crucially on one another, a capability that has demonstrably improved machine translation
by making sure that the translated output represents all of the given input 
even when their word-orders do not correspond exactly \cite{vaswani2017attention}. This has led to rapid advances in the field,
including the contextualized BERT \cite{devlin2018bert} and ELMo \cite{peters2018deep} models. For example, the ELMo model reports
good results on traditional NLP tasks including question answering, 
coreference resolution, semantic role labelling, and part-of-speech tagging, 
and the authors attribute this success to the model's different neural-network
layers implicitly representing several different kinds of linguistic structure.
The survey and experiments of \citeA{hupkes2020compositionality} evaluate three such neural networks on 
a range of tasks related to composition, 
concluding that each network has strengths and weaknesses, that the results are a stepping stone rather than and endpoint, 
and that developing consensus around how such tasks should be designed, tested and shared is a crucial task in itself. 

Even without the cost of encoding order information, \citeA{iyyer2015deep} 
demonstrated a deep {\it averaging} network, showing that network depth could
in come cases compensate for the lack of syntactic sophistication in unordered 
models --- with considerable computational savings. At the time of writing,
this has developed into a very open research question: do neural networks need 
extra linguistic information as inputs to properly understand language, or
can they actually {\it recover} such information as a byproduct of 
training on raw text input? Does a complete NLP system need components for
tokenization, part-of-speech tagging, syntactic parsing, named entity recognition, and so on, 
or can some or all of these be replaced by a single vector language model? If so,
how do we describe the way meaningful units are composed into larger meaningful structures
in such a model?

Tensor networks are one of the possible mathematical answers to this question. 
Their use in libraries such as TensorFlow has become ubiquitous (see \cite{geron2019hands}
and numerous papers and packages referenced therein), though as cautioned above,
users should be wary of the equivalence between the use of matrices and multidimensional arrays 
and tensor algebra that is often assumed in software documentation.
More explicit evidence is presented by work that continues to build upon Smolensky's introduction
of tensors to AI: for example \citeA{mccoy2020tensor} present evidence that the sequence-composition
effects of Recurrent Neural Networks (RNNs) can be approximated by Tensor Product Decomposition Networks, 
at least in cases where using this structure provides measurable benefits over bag-of-words models
(see also \citeA{mccoy2018rnns} for a more detailed presentation). It has also been shown that Tensor Product Networks
can encode grammatical structure more effectively than LSTMs for generating image captions \cite{huang2017tensor}, 
\dom{achieving for example a BLEU-4 score
of 0.305 compared the CNN-LSTM's result of 0.292}.
Tensor product networks have also been used to construct an attention mechanism from which grammatical structure can be recovered
by unbinding role-filler tensor compositions \cite{huang2019attentive}. Explicitly quantum
networks for natural language processing are described by \citeA{wiebe2019quantum}. 
The range of challenges and application opportunities in AI for compositional vector representations is 
by now much-studied and valued, and quantum-inspired tensor networks already successfully combine and extend 
many of the mathematical features that are core to AI today.

%%%
\section{Physical Implementations on Quantum Computers}
\label{sec_implementation}

Quantum computing has regularly made the front page of scientific news 
from 2019 to 2021. This section gives a glimpse of opportunities this opens for quantum mathematics in AI.

With AI based on classical computation, vectors and matrices are already ubiquitous. 
While the algebra and geometry behind them contains many riches that overlap with quantum mechanics, 
studying this overlap in detail is an investment that many researchers might consider esoteric and risky --- 
there are more mainstream state-of-the-art ways to make dramatic progress in AI and machine learning without 
learning quantum theory and tensor algebra. But what if there was a platform for computation where 
vectors could be represented in exponentially smaller memory, and instead of being an
operation with quadratic cost, the tensor product was just the most suitable mathematical representation of a natural system?
Quantum computers may soon provide this.

As outlined in Section \ref{sec:tensor_qm}, the tensor product arises naturally in quantum mechanics. 
Computations using quantum circuits can be used to manipulate tensor products (for example, from separable 
to entangled states \cite[Ch 4]{bernhardt2019quantum}), but no `work' is required to compute the tensor product
of a 2-qubit system in the first place: the wavefunction or state of a 2-qubit system {\it is} represented as the tensor product
of the individual qubit systems, irrespective of any computational operation we perform on these qubits \cite[\S 2]{ying:quantum}.

Perhaps the biggest computational promise that follows from the tensor product
is the potential for representing exponentially more information. A classical register of $n$ physical bits
can represent $n$ bits of information (choices between 0 and 1). 
In quantum computing, these combinations correspond to basis vectors ---
if the state vector of a qubit is represented as a vector in $\mathbb{C}^2$ with basis
vectors $\{\ket{0}, \ket{1}\}$, then a register with $n$ qubits is represented as
a state vector in $\bigotimes_1^n \mathbb{C}^2$, which has dimension $2^n$. 
Instead of a state of the whole system, each of those $2^n$ combinations represents a basis vector
for a coordinate, and if each of these coordinates can be accessed,
the capacity of the memory grows exponentially instead of linearly in the number of qubits in the register!
In AI applications, it is easy to see the appeal of this --- an embedding vector of 256 dimensions using a 4-byte floating point
number for each coordinate requires 1KB of RAM, which can in theory be represented by a tensor product of 8 qubits.
Of course, there are huge challenges with this idea, starting with the physical fact that we could never observe 
such continuous coordinates directly, only their probabilistic quantization to a pure state upon measurement. Nonetheless,
the promise of exponential quantum memory has led to ingenious research ideas (particularly the `bucket-brigade'
qRAM protocol of \cite{giovannetti2008quantum}), and the idea is even presented in an early-reader board book for children \cite{ferrie2018quantum}. There are still practical physical challenges and theoretical caveats to any 
premature claims that `quantum algorithms give exponential improvements' \cite{aaronson2015read}.
Still, if the promise of exponential qRAM is obvious and compelling enough to excite the youngest readers, 
it is hard to imagine that no breakthroughs will happen in the next few decades. 

Still, by the end of 2019, most of the papers published on quantum algorithms were, from a physical point-of-view,
advanced thought-experiments or simulations: they present mathematical techniques or simulated results on classical hardware,
not experimental results from quantum computers. Shor's quantum algorithm for prime factorization was published in 1994 \cite[Ch 9]{bernhardt2019quantum}, 
but at the time of writing, the largest integer factorized on a quantum computer appears to be 291,311 \cite{li2017high}.
There are very particular problems that can already be solved much faster on a 
quantum computer than a classical computer, including the problem used by \cite{arute2019quantum}
which was to simulate and predict the output of a pseudo-random quantum circuit. However, 
this task was proposed {\it in order to} demonstrate quantum advantages, rather than for some existing use-case,
a critique expounded by \citeA{zhong2020quantum}. 

% The term `quantum supremacy' was used for this result, but controversially --- 
% it is relatively unsurprising that simulating a quantum computer can be done faster on a quantum than a classical computer,
% and the claim that this demonstrates `supremacy' for quantum computers
% has been criticized as misleading and potentially harmful \cite{pednault2019quantum}. The term `quantum computational advantage'
% is sometimes recommended as a more guarded alternative by authors including \cite{zhong2020quantum}, whose work 
% provided a second example of such computational speedup by demonstrating a Boson sampling task that ran
% an estimated $10^{14}$ times faster than would be possible on a classical supercomputer. These authors also point out that
% this task was proposed in order to demonstrate quantum advantages --- it is not in itself a problem that currently needs solving in AI. 

There are several research developments
that {\it are} poised to take advantage of quantum computers in AI, in areas including deep learning in particular \cite{wiebe2014quantum},
machine learning in general \cite{biamonte2017quantum}, and language processing including parsing \cite{wiebe2019quantum}, but 
it remains normal for such papers to develop algorithms and sometimes results on real datasets without actually implementing
them on quantum machines. Even papers that describe physical implementation in detail including those in 
quantum optical neural networks \cite{steinbrecher2019quantum} and single photon image classification
\cite{fischbacher2020single} report results on high-end classical hardware and proposals for quantum hardware. 
This is not a negative criticism, just a description of the research frontier:
in celestial mechanics, the orbit of a satellite was studied and predicted for centuries before
spacecraft could actually be launched.

In 2020 and 2021, this has been changing: AI experiments on quantum hardware have been successfully carried out. 
The single-photon classification experiment of \citeA{wang2020quantum} uses a quantum-mechanical platform with a single qubit.
For comparison, the simulated work of \citeA{fischbacher2020single} classifies all ten digits, whereas \citeA{wang2020quantum}
build a classifier that only distinguishes zeros and ones: this exemplifies the sort of tradeoff researchers have made
in using real-but-limited quantum computing resources. \dom{In the work recently reported by \citeA{abbas2021power},
an actual quantum neural network is trained and shown to learn faster and more effectively than a classical network
(as measured by Fisher information and effective dimension), showing as much as a 250\% improvement 
over classical training using the \texttt{ibmq\_montreal} 27-qubit hardware.}

In natural language processing on quantum computers, the most dramatic development so far is perhaps the demonstration of a working
Quantum NLP system on one of IBM’s quantum devices \cite{meichanetzidis2020semspace,coecke2020foundations}. 
Mathematically speaking, this work has several key ingredients. A key part is the use of compact closed categories 
as an axiomatic model for quantum mechanics \cite{abramsky2009categorical}, which enables the DisCoCat model from \citeA{coecke2010mathematical}
to use the same mathematical language. Diagrammatic calculus, developed over some years and explained thoroughly by \citeA{coecke2017picturing},
enables quantum structures for sentences that combine their syntactic structure (in terms of combinations) and semantic content (in terms of vectors)
to be represented together. And the paper of \citeA{meichanetzidis2020semspace} explains in detail how this enables a quantum compiler
to break these expressions down into quantum gates, compiled into quantum circuits and run on a quantum computer. The sentences used to train the model
are still simple toy examples such as {\it `Alice loves Bob'} and {\it `Bob who is cute loves Alice who is rich'} --- as with prime factorization and
image recognition, the implementation using qubits in quantum computers does not yet rival the scale of NLP on classical hardware --- but
the first implementation of a compositional NLP system on quantum hardware has been accomplished successfully, and given the level of
investment in both quantum computing and NLP, more will follow.

\section{Other Areas Related to AI and Quantum Theory}

This final section refers briefly to other areas related to both AI and quantum theory that, to avoid excessive length,
have not been emphasized in this paper, but which are likely to be important avenues for future work. 
Many more papers can be found on this stream of topics than we are able to list here. We encourage interested readers to investigate the references provided in this section for further links to other papers and more detailed avenues. 

\subsection{Quantum Search and Automated Problem Solving}

Quantum search is a large algorithmic topic in itself. In this context, we mean `search' 
in the sense of `database search' or `tree search', rather than the search engines of information retrieval: quantum search
algorithms are typically designed to find a particular element that solves a given problem in computationally faster time than classical 
algorithms \cite[Ch 6]{nielsen2002quantum}. A canonical example of such a speedup is Grover's search algorithm, which 
locates a unique element out of $n$ choices in $\sqrt{n}$ time \cite[Ch 9]{bernhardt2019quantum}. The details are not covered here,
though the process itself is a hallmark example of a quantum oracle that manipulates bases step-by-step in such a way that the 
states of the non-solutions cancel one another out and the state of the solution eventually `sticks out' after at most $\sqrt{n}$ repetitions.

Grover's search algorithm has been used as a building block for approaching other higher-level problems.
These include various forms of tree 
search \cite{tarrataca2011tree} and challenges that can be formulated as symbol manipulation problems, such as a block puzzle 
\cite{tarrataca2011problem}. For an overview see \citeA[Ch 10--12]{wichert:principles}. It is worth noting that 
core data structure operations such as tree search have been investigated for basic hardware components such as 
qRAM \cite{giovannetti2008quantum} while they are also being developed for higher-level problem solving in quantum AI: 
for example, \citeA{arunachalam2015robustness} analyse the importance of reducing the number of lookup 
operations in the `bucket brigade' qRAM protocol error-rates (demonstrating in the process some of the great difficulties
in implementing a real qRAM).

This evolution is historically different from the development of classical computing, in which
basic data structures were reliably and readily available decades before the heyday of machine learning. (This is not
to say that there are no more developments to be made in classical data structures, but that the machine learning
practitioner often has no reason to wonder whether a dictionary lookup is from a binary search tree vs. a hashmap --- 
they `just work'.) By contrast, in quantum computing, these innovations are happening at the same time and facing similar 
challenges --- quantum heuristics are sometimes needed not only to expedite computation, but to
give {\it more} reliable results, because reducing the number of operations reduces errors.
While such details might be seen as frustratingly low-level in most of machine learning today, this is an area where
quantum computing and quantum AI may cross-fertilize each other's early development.

\subsection{Quantum Probability and Cognition}

The geometric and algebraic structures of quantum theory affect probability as well as logic. In quantum theory, the probability
of finding a system in state $\ket{\psi}$ to be in the pure basis state $\ket{\lambda_i}$ is equal to $|\braket{\lambda_i|\psi}|^2$,
which is called the Born rule. For normalized unit vectors, the scalar product $\braket{\lambda_i|\psi}$ is well-known to be the 
cosine of the angle between the two vectors --- so in quantum theory, probability is geometrically related to angles between vectors.
This contrasts with so-called classical probability where probabilities are obtained from ratios between volumes \cite[Ch 2]{isham:quantum}\cite[Ch 2]{khrennikov-ubiquitous}. Quantum probability behaves differently from classical probability: for example, conditional probability depends on the 
spectrum of the observable in question \cite[p. 35]{khrennikov-ubiquitous}, and the fact that a measurement of one of the variables was performed beforehand 
\cite[p. 165]{isham:quantum}.

Quantum probability has been used to explain and 
accurately model order-effects on attitudes, where asking people the same questions in a different order has been shown to give different answers \cite[Ch 3]{busemeyer.bruza:quantum}, resulting in various disjunction effects in decision making. For example,  preferences made with more information have been shown to be different
from those made with less information, irrespective of what information has been learned in the meantime, a violation of the Sure Thing Principle 
\cite{pothos.busemeyer:can}. In semantics, such models have been used to model compositional behavior including nonseparability in the human 
lexical representations \cite{bruza2015probabilistic}.

The affordances of quantum probabilities in cognitive modelling have been exploited by some authors to support more human-like models of automated reasoning and decision making. 
For example, \citeA{kitto.boschetti:attitudes} made use of the structural nature of quantum probability in an agent based model capable of representing phenomena such as cognitive dissonance and attitude change in a social context.
Similarly, a stream of work in developing quantum-like Bayesian network models \cite{moreira.tiwari.ea:quantum} has resulted in sophisticated methods for modelling the evolution of the beliefs of a decision maker.

\subsection{Quantum Probability and Language Modelling}

A key part of the language modeling work of \citeA{sordoni_modeling_2013} was the use of off-diagonal elements in
a matrix to model the joint probability of bigrams (where the on-diagonal elements model unigram probabilities). 
The mathematics of this particular construction is presented intuitively and explored thoroughly in the 
{\it statistical algebra} of \citeA{bradley2020interface}. This leads to a surprising and clear
correspondence between Formal Concept Analysis \cite{ganter-formal}, in which objects and attributes 
are represented in a lattice, and quantum probability. The basic construction is to take the cross table (the matrix showing
which objects possess which attributes), and multiply it by its transpose, which gives a matrix whose off-diagonal elements
capture the overlaps between objects. This conditional dependence information is reflected in the partial trace, which is a 
basis-dependent mapping from the product space $V^\star \otimes V$ back to $V$.
\citeA{bradley2020interface} argues that the partial trace is the quantum analogy of the classical process of marginalizing a joint
distribution --- but whereas in classical probability, marginal probabilities retain no information about the 
overlaps in the original joint distribution, in the quantum case, some of this information is preserved.
This richness may lead to quantum probability becoming a more common tool of choice for language and concept models.

%%
\begin{comment}
\dom{\subsection{Adiabatic Quantum Computing}
This might be a very short section based on \cite[Ch 16]{wichert:principles} to give an example of where
the quantum maths and quantum AI are literally / physically identical.
}
\end{comment}

%%
\subsection{Real and Complex Numbers in AI}

One place where physics, classical logic, and machine learning use quite different fundamental mathematical building blocks is in the choice of number field. Machine learning and AI tend to use real vectors because the basic features are measurements; classical logic and computer science have used binary numbers thanks largely to the enormous influence of George \citeA[Ch 2]{boole-laws}; whereas physics most prevalently uses complex vectors, not just in quantum mechanics but also in electrodynamics.

So far, the reasons for {\it not} using complex and binary numbers in machine learning seem to be twofold (and reasonable): there are no widespread and immediately intuitive interpretations for them; and the field is progressing very quickly without them. However, there are computational benefits to using complex numbers. For example, the convolution operator, sometimes used for vector binding, is just the addition of phase angles, and is thus $O(n)$ rather than the $O(n \log n)$ of the real convolution operation optimized using Fast Fourier Transforms (see \citeA{plate-hrr}).
And in several experiments on automatic inference, we have ourselves found that using binary and complex vectors sometimes yields much better results than their real-valued counterparts, for reasons not yet properly understood \cite{widdows2015reasoning}. 
There are a few machine learning papers on using complex
numbers and even quaternions and octonions in machine learning, for example \citeA{trabelsi2018deep}. As quantum computing becomes more mainstream, 
we should expect the use of complex numbers to become more widespread.

\subsection{Complex Time Evolution}

Imaginary and complex numbers are particularly important in quantum mechanics for representing the momentum operator $\hat{p} = -i \hbar \frac{\partial}{\partial x}$
and the resulting Hamiltonian in the Schr{\"o}dinger wave equation. This technique has roots in harmonic analysis, particularly with the
simple form of the plane wavefunction $\Psi(x, t) = A e^{i(kx - \omega t)}$. Harmonic functions can be written in this way thanks to the identity $e^{i\theta} = 
\cos{\theta} + i \sin{\theta}$, and the technique for representing time evolution could potentially be applied to situations where 
harmonic functions are used to model word position as attention moves along a sentence \cite{vaswani2017attention}.

\subsection{Again, Why Quantum Theory?}
\label{sec:again-why}

Throughout this paper we have introduced and surveyed several areas where mathematics used in quantum theory is also fruitful in AI. However, we are yet to answer the questions
``What is quantum mathematics?'' and ``Why is it important in AI?'', delaying them to this point. This has enabled us to demonstrate the many and diverse ways in which mathematics has been abstracted from quantum theory and applied to AI. But there is no  clear characterization of what precisely makes mathematics `quantum'. 

This is well-illustrated by contrast with 
the other great pillar of 20th century physics, general relativity. The mathematics of general relativity can be defined crisply as 
the differential geometry of Lorentzian manifolds, which are smooth 4-manifolds with a pseudo-Riemannian metric of signature (1, 3) --- that
is, there must be 4 dimensions, and the metric must treat three of them interchangeably (the spacelike dimensions) and one with
the opposite sign (the timelike dimension), 
giving a local distance metric that takes the form $dt^2 - dx^2 - dy^2 - dz^2$ (choosing units so that the speed of light is 1). 

After years of searching, we are confident that {\it there is no similarly clear characterization of quantum mathematics}.
That disjunctions are non-distributive or products non-separable have been proposed, but can be rejected on the grounds that such
mathematical structures are prominent in many other areas, and many quantum models do not explicitly use these properties. 

So the question ``What makes mathematics distinctly quantum?'' remains a matter for discussion. In our opinion, 
quantum mathematics as a whole is the mathematics of self-adjoint operators in complex Hilbert space. In theory,
complex numbers are needed to guarantee that roots of polynomials exist, which guarantees that eigenvalues and eigenvectors exist.
The self-adjoint requirement guarantees that these eigenvalues are real numbers, which is necessary for them to represent physical properties.
This mathematical argument is standard: in this case, we are following particularly the narrative of \citeA[Ch 6]{coecke2017picturing}, a
similar presentation is given by \citeA{rijsbergen-geometry}, and indeed \citeA[Ch 1, 2]{dirac-quantum}. 

Much of this mathematical apparatus is general, and has been used in a wide array of fields beyond quantum theory. 
In this paper we have demonstrated a number of these overlaps with AI. However, with this point comes an important caveat: 
most AI applications are only using a part of quantum mathematics, since few use complex numbers.
For example, the information retrieval applications discussed in Section \ref{sec:quantum-ir} are explained explicitly in terms of quantum logic 
(in the case of \citeA{widdows-quantum}) and quantum probability (in the case of \citeA{sordoni_modeling_2013}), but use only real numbers throughout.
Rather than a criticism, this could become a recommendation: quantum-inspired AI systems should systematically consider and investigate the 
use of complex numbers.

Instead of embodying a distinct piece of mathematics, the motivation for developing quantum-inspired models 
in AI and related areas has sometimes been the obvious shortcomings of so-called `classical' models. 
Quantum theory acknowledges that the future is not wholly predictable, that combined systems cannot always be 
separated into atomic parts, and that the order in which observations are made sometimes affects the outcome. 
These points are often obscured by the oft-repeated claim ``classical physics is the 
physics of everyday life'', even though {\it quantum physics is more like daily life}
in these respects (points discussed in more detail by \citeA{busemeyer.bruza:quantum} and 
\citeA{widdows2007quantum}). As these areas have been explored in detail, with hindsight it
is easy to understand why quantum structures were seen as promising alternatives to 
reductionist symbolic determinism: and once this approach is considered, mathematical 
overlaps between quantum mechanics and other application areas are easy to find. 
Without the advent of quantum computing, these overlaps may have been considered a 
mathematical curiosity: a good motivation perhaps
for individual system designs so long as they perform well in practice. But now with
quantum computing coming of age, the importance of quantum mathematics in computer science 
is set to increase, because it will be key to answering the question ``What's the most
effective way to implement this?''

\section{Conclusion}

Quantum theory and artificial intelligence use much of the same mathematics, and the importance of the overlapping areas 
has grown with the ubiquitous use of vectors in AI. As well as simple superposition and similarity operations, 
quantum theory offers well-studied formulations of logical inference, concept combination, and probability, many of which
have been demonstrated to give good results on various AI tasks, including examples from natural language processing, image processing,
reasoning and inference. The mathematics used in these applications is often well-developed thanks to its use in other 
fields, and in particular, Grassmannian and tensor algebra offers more in terms of duality and symmetry than is typically used in AI.
Some of these algebraic operations are costly on classical hardware, but much cheaper or even free in quantum computation, which
in 2021 is rapidly becoming more practical. The authors hope that this paper helps AI researchers familiar with vectors 
and their uses to turn more often to quantum theory as a rich source of mathematical motivation for advances today, and
to approach the advent of quantum computing with confidence and excitement.

%%%
\section*{Appendix: Category Theory for Quantum AI}
\label{sec:category_theory}

Category theory is one of the less familiar branches of mathematics
to many researchers in AI, and this appendix is designed as a very brief
supplement. Interested readers are referred particularly to the survey article of \citeA{selinger2010survey} and the 
graduate text of \citeA{heunen2019categories} for a proper explanation.

Much of modern mathematics can be thought of as the study of objects
with some defined structure, along with maps that preserve this structures.  
Standard examples well-known by the middle of the 20th century include
sets and maps, groups and homomorphisms, topological spaces and continuous maps,
differentiable manifolds and differentiable maps, and of course,
vector spaces and linear maps.

The abstract notion of a category was developed in the 20th century to reason about such
structures in a common fashion. A category consists of a collection of {\it objects}, and for
every pair of objects $A$ and $B$, a collection of {\it morphisms} from $A$ to $B$. Morphisms must
be composable in sequence in an associative manner (so that $f\circ(g\circ h) = (f\circ g)\circ h$), 
and there must be an identity morphism $\mathrm{id}$ so that $\mathrm{id}\circ f = f\circ \mathrm{id} = f$. 

In many categories, the morphisms are some kind of function. These include {\bf Set},
the category of sets and maps, and {\bf Vect}, the category of vector spaces and 
linear maps. 
There are also important cases where the morphisms are not functions,
including the category of sets and {\it relations}, which is often called {\bf Rel}.
A relation between two sets $A$ and $B$ is a subset of their Cartesian product $A\times B$, intuitively 
a `many to many correspondence', and relations can also be composed in a way that satisfies the definition of a category 
\cite[\S 0.1.3]{heunen2019categories}. Because of the potential for one-to-many correspondences, 
if we think of `applying a morphism' as `evolving in time', the category {\bf Rel} can be used as a model for
`nondeterministic classical physics', because a single state representing the present can evolve into more than one
possible state representing the future.

Following the pattern of abstract algebra, further definitions are introduced and their
properties and consequences explored. Some of these (here described intuitively but not precisely) are as follows.
A {\it functor} is itself a mapping between two categories that maps objects to objects and morphisms to morphisms in a regular composable fashion.
The {\it opposite} or {\it dual}
category $C^{op}$ has the same objects as $C$ but `reversed' morphisms: if we define the set of morphisms from $A$ to $B$ in the category $C$ as 
$C(A, B)$, then the set of morphisms from $B$ to $A$ in $C^{op}$ is defined
by $C^{op}(B, A) = C(A, B)$. This is simple but initially confusing, especially with the 
most obvious category {\bf Set} of sets and functions --- how can a function from $A$ to $B$ be considered
a morphism from $B$ to $A$? In this surprising case, the answer is just `the definition
satisfies identity and associativity, so it forms a category'. However, with the categories of {\bf Rel} of sets and relations, 
and {\bf FVect} of (finite dimensional) vector spaces and linear maps, the notion of dual categories becomes
much more concrete, as the objects and morphisms themselves have duals,
not only the category as a whole. Much of the power in categorical quantum mechanics arises from the use of category theory to explore the properties
and consequences of duality in these settings.

A {\it closed} category is one where the morphisms themselves have the structure
of objects in the category. For example, the mappings from a finite set $A$ to a finite set $B$ themselves form a finite set 
(an element of $B$ is selected for each element of $A$, so the number of such possibilities is $|B|^{|A|}$), so
finite sets are a closed category. 
A {\it monoidal} category is one where objects from the same category can be composed with 
one another `in parallel' using an operator written $\otimes$ (as well as morphisms being composed `in sequence'), and there
needs to be an identity object $I$ that satisfies basic compatibility conditions with the identity morphism. 
This gives the objects of the category the structure of a monoid, 
which is a group without the requirement of each element having an inverse.
In a monoidal category, an object $A$ may have a {\it dual} object $A^\star$, with canonical morphisms from $I$ to $A^\star \otimes A$ and vice versa.
If these morphisms contract to the identity in given ways, the category is called {\it compact}. 

%(The term is related to the definition of a compact space in topology by the representation theory of compact Lie groups: in fairness, this gives an indication of
%how much background in abstract algebra and topology is involved in the motivations
%and understanding of category theory.)

A {\it dagger} category is one where each morphism $f: A \rightarrow B$ has a corresponding dagger
morphism $f^\dagger: B\rightarrow A$, with algebraic properties generalized from those of the adjoint operator in linear algebra.
Dagger categories were used by \citeA{abramsky2004categorical}
for describing quantum information protocols,
using the term {\it strongly} compact closed categories. The term `dagger category' was introduced as part of a more general exploration of their
mathematical properties by \citeA{selinger2007dagger}. 

The connection of notation and terminology with linear algebra is quite deliberate: 
finite dimensional complex vector spaces form a monoidal category whose
monoidal product is the tensor product, and whose dual objects are the dual vector spaces.
Hence the notation $A^\star$ for the dual of $A$, and $A\otimes B$ for the monoidal category 
product, are just the same definitions as given in linear algebra.

The relationship with quantum mechanics, called {\it categorical
quantum mechanics}, is due particularly to \citeA{abramsky2009categorical}, and has become part of the mathematical backbone of work on 
understanding quantum processes by 
\citeA{coecke2017picturing} and in related works. Categorical quantum mechanics seeks to
describe quantum mechanics and quantum processes in terms of mathematical properties like those outlined
above.

This contributes to quantum mathematics in two ways. First, the results of quantum
mechanics can be linked even more directly to key mathematical concepts. For example,
\citeA[Ch. 4]{heunen2019categories} uses the constrasting monoidal structures of 
the categories {\bf Set} and {\bf FHilb} (finite dimensional Hilbert Spaces and bounded linear maps) to explain why we should expect that state can be
uniformly copied in classical computing but not in quantum computing; and
they
relate categorical properties to aspects of entanglement and quantum teleportation throughout. 
Second, understanding such phenomena 
at a more abstract mathematical level can lead to applications in other areas and 
cross-fertilization. For example, 
the category of sets and relations (more general than the category of sets and maps, 
because a relation can be many-to-many) sometimes behaves much more like the category
of vector spaces than that of sets, particularly in relation to entanglement,
which the statistical algebra of \citeA{bradley2020interface} uses to describe
concepts in a formal concept lattice. And of particular importance, the
use of category theory links the {\it pregroup grammars} of Lambek calculus to 
the composed vector representations of \citeA{coecke2010mathematical},
the algebraic representation of pronouns \cite{sadrzadeh2013frobenius,sadrzadeh2014frobenius2}, 
and the 
gate-based quantum circuit implementation of \citeA{meichanetzidis2020semspace}.
Here, for example, the notion of monoidal products and duals is used to give a categorical 
model for a linguistic notion such as ``a transitive verb operates on two nounphrases'' 
and a vector notion such as ``a type (0, 2) tensor operates on two type (1, 0) vectors'', in such
a way that the verb and noun representations can be implemented on a quantum computer.

%Admittedly, the technical material involved in using category theory in these ways is 
%challenging: readers unfamiliar with this field are likely to find the works
%cited in this section difficult. We hope that the summary and connections here
%are enough to arouse interest and give readers a sense that an investment in learning
%about category theory may be rewarding.

While category theory can sound very abstract and technical, it gives clearer insights into
why apparently similar systems sometimes behave differently, and why some apparently very different
systems share common structures. Today especially, it provides an ideal opportunity for mathematicians
with a background in algebra to contribute to technologies for computing, security, and communication.

\bibliography{QC,infomap}
\bibliographystyle{theapa}

\end{document}